\documentclass[journal]{IEEEtran}

\usepackage{cite}
\usepackage{graphicx}
\usepackage{subfigure}
\usepackage{amsmath}
\usepackage{amssymb}
\usepackage{tabularx}
\usepackage[table,xcdraw]{xcolor}
\usepackage{multirow}
\usepackage{color}
\usepackage{algorithm}
\usepackage{booktabs}
\usepackage{hhline}
\usepackage{lineno}
\usepackage{colortbl}
\usepackage{amsthm}
\usepackage[colorlinks]{hyperref}
\usepackage[]{algpseudocode}
\usepackage[english]{babel}
\usepackage[utf8]{inputenc}
\usepackage[super]{nth}

\newcommand{\RN}[1]{\textup{\uppercase\expandafter{\romannumeral#1}}}

\usepackage[T1]{fontenc}

\makeatletter
\def\hlinewd#1{%
	\noalign{\ifnum0=`}\fi\hrule \@height #1 \futurelet
	\reserved@a\@xhline}

\definecolor{dbcolor}{rgb}{0,0,1}

\definecolor{jhcolor}{rgb}{1,0,0}

\begin{document}
	
	\title{Deep Monocular Depth Estimation Leveraging a Large-scale Outdoor Stereo Dataset}

\author{Jaehoon Cho,~\IEEEmembership{Student Member,~IEEE,}
	Dongbo Min{$^\dagger$},~\IEEEmembership{Senior Member,~IEEE,}\\
	Youngjung Kim,~\IEEEmembership{Member,~IEEE,}
	and Kwanghoon Sohn,~\IEEEmembership{Senior Member,~IEEE}
	\thanks{J. Cho, and K. Sohn are with the School of Electrical and Electronic Engineering, Yonsei University, Seoul 120-749, South Korea (e-mail: \{rehoon,khsohn\}@yonsei.ac.kr).}%
	\thanks{D. Min is with the Department of Computer Science and Engineering, Ewha Womans University, Seoul 03-760, South Korea (e-mail: dbmin@ewha.ac.kr).}%
	\thanks{Y. Kim is with the Agency for Defense Development, Daejeon 34186, South
		Korea (e-mail: read12300@add.re.kr).}
	\thanks{$\dagger$ Corresponding author}}%

	\markboth{}%
	{Shell \MakeLowercase{\textit{et al.}}: Bare Demo of IEEEtran.cls
		for Journals}

	\maketitle
	\IEEEpeerreviewmaketitle
	
	\begin{abstract}

	Current self-supervised methods for monocular depth estimation are largely based on deeply nested convolutional networks that leverage stereo image pairs or monocular sequences during training phase.
	However, they often exhibit inaccurate results around occluded regions and depth boundaries.
In this paper, we present a simple yet effective approach for monocular depth estimation using stereo image pairs.
The study aims to propose a student-teacher strategy in which a shallow student network is trained with the auxiliary information obtained from a deeper and more accurate teacher network.
Specifically, we first train the \textit{stereo} teacher network by fully utilizing the binocular perception of 3-D geometry, and then use the depth predictions of the teacher network to train the student network for monocular depth inference.
This enables us to exploit all available depth data from massive unlabeled stereo pairs.
We propose a strategy that involves the use of a data ensemble to merge the multiple depth predictions of the teacher network to improve the training samples by collecting non-trivial knowledge beyond a single prediction.
To refine the inaccurate depth estimation that is used when training the student network, we further propose stereo confidence guided regression loss that handles the unreliable pseudo depth values in occlusion, texture-less region, and repetitive pattern.
To complement the existing dataset comprising outdoor driving scenes, we built a novel large-scale dataset consisting of one million outdoor stereo images taken using hand-held stereo cameras.
Finally, we demonstrate that the monocular depth estimation network provides feature representations that are suitable for high-level vision tasks.
The experimental results for various outdoor scenarios demonstrate the effectiveness and flexibility of our approach, which outperforms state-of-the-art approaches.

	\end{abstract}


	\section{Introduction}
		\vspace{-3pt}
	\IEEEPARstart{O}{btaining} 3-D depth of a scene is essential to alleviate a number of challenges in robotics and computer vision tasks including 3-D reconstruction~\cite{yang2018dense}, autonomous driving~\cite{cao2015perception,choi2018kaist,cho2018multi}, intrinsic image decomposition~\cite{kim2016unified}, and scene understanding~\cite{gupta2015indoor}. 
	The human visual system (HVS) can understand the 3-D structure by perceiving the depth value of the scene using binocular fusion.
	Such a mechanism has been widely adopted in computational stereo approaches that establish correspondence maps across two (or more) images taken of the same scene~\cite{scharstein2002taxonomy}. 
	This approach has achieved outstanding performance in recent studies~\cite{han2015matchnet,MCCNN,luo2016efficient,DispNet,kendall2017end,chang2018pyramid}.
	Furthermore, numerous monocular depth estimation approaches have been developed based on monocular cues,  for example, inclusion of the object contour~\cite{lee2009geometric}, and segmentation~\cite{hoiem2005geometric}.
	However, most methods rely heavily on handcrafted rules based on one or a few monocular cues, and thus they often fail to capture plausible depth from a single image and are effective only at very restricted environments.

	Recently, owing to advances in deep neural networks, depth prediction from a single image has advanced considerably with the aid of convolutional neural networks (CNNs)~\cite{Eigen2015,NYU,Liu2015,monodepth17,semi17,SVS18,godard2019digging}.
	However, supervised learning approaches for monocular depth estimation necessarily have several limitations outdoors.
	Monocular depth estimation needs (semi-)dense depth maps as pixel-level supervision for training the deep network, and constructing such large-scale training data with depth maps is extremely challenging.
	An active depth sensor, LiDAR, is commonly used to acquire depth maps, however, it is usually of low resolution and very sparse, for example, less than 6\% in the KITTI dataset~\cite{KITTI}.
	Owing to its sparsity, it cannot cover all salient objects in a scene.
	Additionally, the sensing device is very expensive and is often adversely affected by several internal degradations, such as imperfect sensor calibration and photometric distortions.
	Thus, most of the existing public datasets provide only a small number of depth maps for rather limited scenes.
	For example, they mostly consist of driving scenes obtained from the depth sensor mounted on a vehicle~\cite{KITTI,Cityscape}.

\begin{figure}[!]
	\centering
	\subfigure{\includegraphics[width=0.49\textwidth]{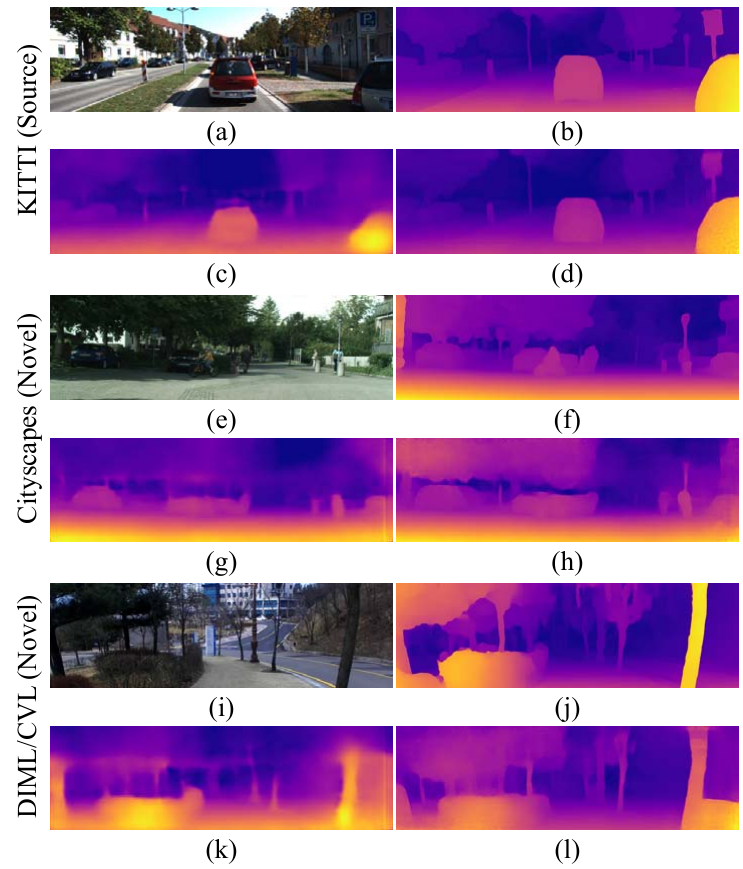}}\vspace{-8pt}
	\\
	\caption{
		Sample images collected from various datasets and estimated depth maps: (a), (e), (i) input images, (b), (f), (j) depth maps predicted by deep stereo matching network~\cite{CRL17}, (c), (g), (k) depth maps obtained using state-of-the-art method~\cite{semi17}, and (d), (h), (l) depth maps obtained using our method.
	}
	\vspace{-20pt}
	\label{fig:1}
\end{figure}

{To address the lack of ground truth depth maps, recent works introduce a self-supervised paradigm ~\cite{monodepth17,garg2016unsupervised,godard2019digging} that uses stereo image pairs during the training phase.
	These methods impose the left-right photometric consistency via view synthesis.
	Since it simply measures on RGB color or hand-crafted feature (e.g., SSIM) spaces,
	they inherently suffer from texture-less region, occlusion, and repeated pattern, resulting in
	blurry depth boundaries and fattening artifacts.
	The method~\cite{semi17} uses both the supervised loss using ground truth depth maps and the self-supervised image reconstruction loss.
	However, it still faces a performance degradation due to the supervised loss when the trained model feeds an image from a novel domain.
	Owing to the self-supervised loss, it yields blurry depth results similar to ~\cite{monodepth17,garg2016unsupervised,godard2019digging}, as shown in Fig.~\ref{fig:1}.}
Another proposed method~\cite{SVS18} first synthesizes the right view using deep networks and then performs stereo matching to produce a final depth map inside the monocular depth network.		
However, this two-step method substantially increases the computational complexity.

In this paper, we present a novel approach based on a student-teacher strategy. 
Our approach obviates the need to use massive ground truth depth maps.
We propose the use of a massive number of stereo image pairs, which are relatively easy to obtain, and a small amount of training data with ground truth depth maps provided in existing public datasets.
First, the teacher network for stereo matching is trained using a small amount of training data with ground truth depth maps.
Subsequently, multiple estimated depth maps are generated on various scales, and then these maps are fused.
This ensemble approach improves the accuracy of the pseudo-ground-truth depth maps by collecting non-trivial knowledge beyond a single prediction.
Given the fused depth maps, the associated stereo confidence maps are generated via a confidence measure network.
The stereo confidence maps encourage the use of only the reliable depth values in the fused depth maps.
With these maps, we propose a stereo confidence-guided regression loss that employs a mask function.
To use diverse stereo image pairs when training the monocular depth networks, we built a new large-scale dataset named the DIML/CVL dataset, by capturing stereo image pairs of various scenes, including parks, brooks, and apartments (Fig.~\ref{fig:7}).
The experimental results demonstrate that our method outperforms the state-of-the-art method, and the DIML/CVL database is complementary to existing outdoor driving scenes~\cite{KITTI,Cityscape}.	
We additionally demonstrate that our pre-trained model for monocular depth prediction can be used as a powerful proxy task for scene understanding tasks.

{
	Recent works ~\cite{guo2018learning,tosi2019learning} are conceptually similar to the proposed method in that they also exploit stereo matching as a rich knowledge for monocular depth estimation.
	In contrast, our method is more effective than previous works by handling unreliable areas such as texture-less regions and occlusions in stereo matching using confidence guided loss.
}

Our main contributions are as follows:
\begin{itemize}	
	
	\item 
	We propose a novel framework for monocular depth estimation based on a student-teacher strategy.
	
	\item 
	We introduce a data ensemble and stereo confidence-guided regression loss to improve the usage of the pseudo-ground-truth.
	
	\item  We introduce a new large-scale RGB-D dataset, named the DIML/CVL dataset, which is complementary to the existing datasets of outdoor driving images.	
	The dataset is publicly available at \url{https://dimlrgbd.github.io/}.
	
	\item We demonstrate that the feature representation of our monocular depth estimation provides rich knowledge for scene understanding tasks.
	
\end{itemize}

\section{Related Work}
In this section, we briefly review and discuss three lines of work that are the most relevant to our study.

\subsection{Stereo matching}
The objective of stereo matching is to find a set of corresponding points between two (or more) images.
The correspondence map is converted into a depth map using stereo calibration parameters.
Early studies based on CNN attempted to measure the similarity between patches of two images.
Han \emph{et al.}~\cite{han2015matchnet} proposed a Siamese network that extracts features from patches followed by a similarity measure.
Zbontar and LeCun~\cite{MCCNN} computed the matching cost using CNNs and applied it to classical stereo matching pipelines consisting of cost aggregation, depth optimization, and post-processing.
Luo \emph{et al.}~\cite{luo2016efficient} proposed computing the matching cost by learning a probability distribution over all depth values and then computing the inner product between two feature maps.
Note that these approaches focused on computing the matching cost using CNNs and the remaining procedures for the stereo matching still rely on conventional handcrafted approaches.

Recent approaches have attempted to predict a depth map in an end-to-end fashion, achieving a substantial performance gain.
Mayer \emph{et al.}~\cite{DispNet} proposed a new method, named DispNet, which uses a series of convolutional layers for cost aggregation and then employs regression to obtain the depth map.
Pang \emph{et al.}~\cite{CRL17} introduced a two-stage network, named cascade residual learning (CRL), which is an extension of DispNet \cite{DispNet}.
The first and second stages are used to calculate the depth maps and their multi-scale residuals, and then the outputs of both stages are combined to form a final depth map.
Kendall \emph{et al.}~\cite{kendall2017end} introduced a new end-to-end approach that performs cost aggregation using a series of 3-D convolutions.
{Chang \emph{et al.}~\cite{chang2018pyramid} incorporated a contextual information through 3-D convolutions using stacked multiple hourglass networks over cost volume.}
The above-mentioned stereo matching approaches can be utilized as the teacher network in our framework.


\subsection{Monocular depth estimation}

{The performance of the monocular depth estimation has been advanced dramatically through the supervised learning approach that uses depth maps acquired from active sensors as ground truth for input images.}
Eigen \emph{et al.}~\cite{Eigen2015} designed a multi-scale deep network that predicts a coarse depth map and then progressively refines the depth map.
Liu \emph{et al.}~\cite{Liu2015} casted the monocular depth estimation into a continuous conditional random field (CRF) learning problem that jointly learns the	unary and pairwise potentials of the CRF in a unified deep CNN framework.
{
	Fu et al.~\cite{fu2018deep} introduced a discrete paradigm that uses an ordinal regression loss to encourage the ordinal competition among depth values in an end-to-end manner.}

{
	To address the lack of massive ground truth depth maps, several approaches have attempted to learn the monocular depth inference networks using image pairs only in a self-supervised manner.
	Using a left-right consistency constraint, Godard \emph{et al.}~\cite{monodepth17} proposed an improved architecture for training the monocular depth estimation using stereo image pairs.
	Zhou \emph{et al.}~\cite{egomotion17} designed a model to jointly estimate depth and camera pose in a self-supervised manner by leveraging temporally adjacent frames acquired by a single moving camera.
	Similarly, Godard \emph{et al.}~\cite{godard2019digging} exploited consecutive frames to estimate both the depth and the camera pose.
	One drawback of these approaches is that reconstruction loss based on image matching is an ill-posed problem on its own.
	To address this problem, Kuznietsov \emph{et al.}~\cite{semi17} exploited both supervised and self-supervised losses with ground truth depth maps and stereo image pairs.
	This approach still suffers from the performance degradation in novel environments, as presented in Fig.~\ref{fig:1}.
	Luo \textit{et al.}~\cite{SVS18} formulated monocular depth estimation with two sub-networks: a view synthesis network and a stereo matching network. 
	They first synthesize stereo pairs from an input image, and then apply the stereo matching network to produce the depth map.
	Zhao \textit{et al.}~\cite{zhao2019geometry} designed a geometry aware symmetric domain adaptation framework using the ground truth depth maps in synthetic data and epipolar geometry in the real data.
}

{
	Recent studies attempted to use pseudo ground truth depth maps obtained by exploiting stereo matching as a rich knowledge.
	Guo \textit{et al.}~\cite{guo2018learning} trained the stereo matching network using a synthetic stereo dataset.
	The above-mentioned methods did not consider performance degradation incurred by unreliable pseudo depth values in occlusion, texture-less region, and repetitive pattern that can not be addressed using stereo matching methods.
	In contrast, our method is more effective than recent studies by handling unreliable values using confidence guided loss.
}

\subsection{Student-teacher strategy}

Models that are deep and wide are known to tend to be more accurate than shallow models and they provide a large capacity~\cite{ba2014deep,hinton2015distilling}.
The student-teacher strategy concentrates on learning a much smaller model (student) from a large deep network (usually referred to as the teacher).
With this method, the shallow network can be as accurate as the deep teacher network~\cite{xu2018training}. 
Ye \emph{et al.}~\cite{ye2019student} introduced a joint scene parsing and depth estimation network with a shallow model using heterogeneous-task deep teacher networks.
Pilzer \emph{et al.}~\cite{pilzer2019refine} designed a refinement process via cycle consistency and distillation strategies for monocular depth estimation.
In our method, the pseudo ground truth depth maps computed from the existing stereo matching network, which acts as a deep teacher network, are used to train the student network for monocular depth inference.

\subsection{Feature learning via pretext task}

{Several approaches have attempted to leverage a pretext task as an alternative form of supervision in some applications where it is difficult to construct massive ground truth data.}
Noroozi and Favaro~\cite{noroozi2016unsupervised} proposed solving jigsaw puzzles for representing object parts and their spatial arrangements,
and then applied it to object classification and detection tasks.
Pathak \emph{et al.}~\cite{Pathak16} proposed an unsupervised image inpainting approach that generates the contents of an arbitrary image region conditioned on its surroundings.
The learned encoder features are applied to object classification/detection and semantic segmentation tasks.
Larsson \emph{et al.}~\cite{larsson2017learning} investigated image colorization as a proxy task in the replacement of ImageNet~\cite{ImageNet}.

These studies have been successfully transferred to various high-level tasks.
In our study, we demonstrate that the network pre-trained for monocular depth prediction is a powerful proxy task for learning feature representations in scene understanding tasks.

	\begin{figure*}[!]
		\centering
		\subfigure{\includegraphics[width=0.95\textwidth]{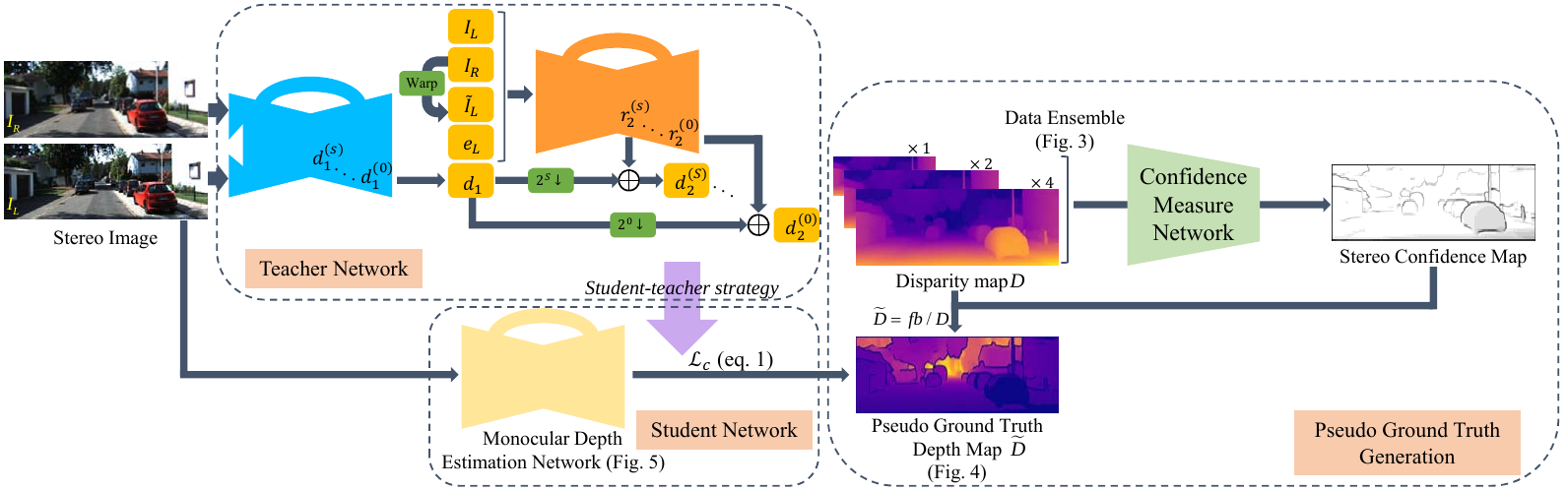}}
		\\
		\caption{
				Proposed framework for monocular depth estimation.
				We first generate depth maps from stereo image pairs using a deep teacher network~\cite{CRL17}. 
				Here, we adopt the data ensemble~\cite{radosavovic2017data} to fuse depth maps estimated on multiple scales. 
				A stereo confidence map~\cite{poggi16} is generated to identify inaccurate estimated stereo depth values. 
				Subsequently, the monocular student network is trained with the pseudo-ground-truth depth maps via stereo confidence guided regression loss.
		}
		\label{fig:2} 
		\vspace{-15pt}
	\end{figure*}

	\section{Proposed Method}

	\subsection{Motivation and overview}

	Owing to the lack of scene diversity, deep networks for monocular depth estimation often undergo performance degradation in novel environments.
	For instance, feeding a single (monocular) image from a source domain, in which the deep networks are trained, yields satisfactory results (Fig.~\ref{fig:1} (c)).
	However, when we test an image from a novel domain, the output depth map produces inaccurate results around occluded regions and depth boundaries.
	The monocular depth estimation network trained with the KITTI dataset~\cite{KITTI} does not work well on the Cityscapes dataset~\cite{Cityscape} and our new dataset.
	In contrast, the stereo matching network using~\cite{CRL17} produces fine-grained depth maps on both the source and novel domains, as shown in Fig.~\ref{fig:1} (b), (f), and (j), even though it is trained with the KITTI dataset only.

	Stereo matching aims to find similar patches from a number of candidates extracted from two images.
	Thus, it is sufficient to train the network with similar patches (positive samples) and dissimilar patches (negative samples)~\cite{MCCNN}.
	The stereo matching network attempts to learn the local patch matching from the cost-volume {(\textit{i.e.,} correlation layer, which explicitly encodes the geometric relationship~\cite{CRL17,DispNet}).}
	Some methods propose training the stereo matching network using two images at once to additionally leverage a global context on stereo matching~\cite{chang2018pyramid}.
	However, the underlying principle is to locally explore the patch-level similarity for two-view matching.
	Contrarily, monocular depth estimation, which infers a depth value from a single image by making use of monocular cues, is highly ill-posed and more challenging than stereo matching.
	Thus, the global context is crucial for predicting the overall 3-D structure of scenes.
	In this regard, the monocular depth estimation network is trained using the image and depth map, rather than a pair of patches extracted from them, to consider the global context.
	This increases the sensitivity of the monocular depth estimation network to the domain difference problem compared with the stereo matching network.
	Thus, a great variety of scenes is needed to train the monocular depth estimation network, whereas the stereo matching network is relatively free from such constraints~\cite{guo2018learning,tosi2019learning}.

	Our overall framework is illustrated in Fig.~\ref{fig:2}.
	We propose a simple yet effective approach for monocular depth estimation by leveraging the student-teacher strategy. 
	The shallow student network learns from the more informative deep teacher network. 
	Our method involves the following steps. 
	Given a number of stereo images, we generate depth maps using the deep stereo matching network trained with ground truth data. 
	When generating depth maps, we fuse depth maps that were estimated on multiple scales to provide non-trivial knowledge from multiple predictions. 
	A stereo confidence map is then estimated as auxiliary data to avoid inaccurate stereo depth values being utilized when training the monocular depth estimation network. 
	The pseudo-ground-truth depth maps are used to supervise the monocular student network via stereo confidence guided regression loss. 
	{The experimental results show that our framework is easily extendable to various domains by leveraging only stereo image pairs.}
	Furthermore, the monocular depth estimation induces feature representation that improves scene-understanding tasks such as semantic segmentation and road detection. 
	
	{
		Differently from other frameworks following a similar work~\cite{guo2018learning,tosi2019learning} that uses existing stereo matching to generate pseudo-ground-truth depth maps and then use these maps for training monocular depth network, we can handle  unreliable areas such as texture-less regions and occlusions in stereo matching using confidence guided loss yielding a notable accuracy improvement compared to the existing solutions.
	}

\begin{figure}
	\centering
	\subfigure{\includegraphics[width=0.45\textwidth]{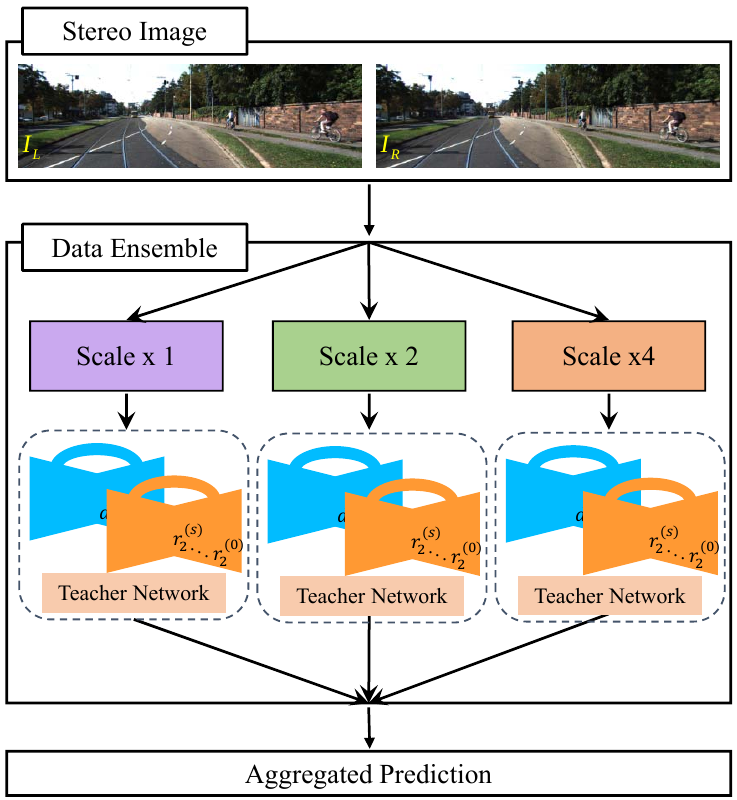}}
	\\
	\vspace{-3pt}
	\caption{Data ensemble approach.
		We generate depth maps of stereo images at different scales via the deep teacher network, and then ensemble them on the smallest scale.}
	\label{fig:3}
	\vspace{-8pt}
\end{figure}

				\subsection{Pseudo-ground-truth generation}	
				
				The stereo matching network takes stereo images $(I_l,I_r)$ as input and outputs a depth map $D$ aligned with the left image $I_l$.
				We adopted cascade residual learning (CRL)~\cite{CRL17} as the stereo teacher network.
				To further improve the depth map of the teacher network, we adopt an ensemble prediction method that merges output depth maps on various scales, as shown in Fig.~\ref{fig:3}.
				It has been shown that the generated data can be improved by applying the same model to multiple transformations (\textit{e.g., } scale, rotation, and flipping) of the input and then aggregating the results~\cite{he2016deep,long2015fully}.
				We generate depth maps on three different scales and average them on the smallest scale.
				The data ensemble can provide auxiliary information from multiple predictions beyond a single prediction.

\begin{figure}
	\centering
	\renewcommand{\thesubfigure}{}
	\subfigure[(a)]{\includegraphics[width=0.235\textwidth,height=0.05\textheight]{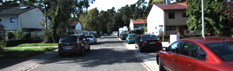}}
	\subfigure[(b)]{\includegraphics[width=0.235\textwidth,height=0.05\textheight]{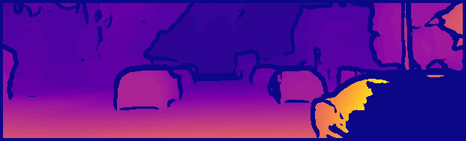}}\\
	\vspace{-8pt}
	\subfigure[(c)]{\includegraphics[width=0.235\textwidth,height=0.05\textheight]{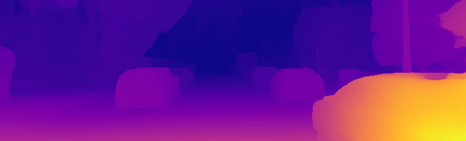}}
	\subfigure[(d)]{\includegraphics[width=0.235\textwidth,height=0.05\textheight]{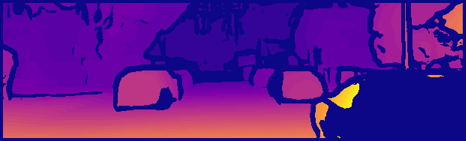}}\\
	\vspace{-8pt}
	\subfigure[(e)]{\includegraphics[width=0.235\textwidth,height=0.05\textheight]{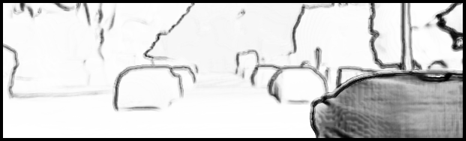}}
	\subfigure[(f)]{\includegraphics[width=0.235\textwidth,height=0.05\textheight]{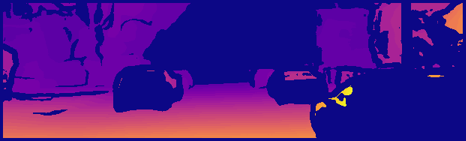}}\\
	\vspace{-3pt}
	\caption{Pseudo ground truth data samples.
		(a) input image, (c) estimated depth map with data ensemble, and (e) predicted stereo confidence map.
		(b), (d), (f): pseudo ground truth depth maps thresholded by the stereo confidence map (e) with = 0.3, 0.55, and 0.75, respectively.							
		Black pixels indicate unreliable pixels detected by the stereo confidence map.} \vspace{-15pt}
	\label{fig:4}		
\end{figure}

\begin{figure}
	\centering
	\renewcommand{\thesubfigure}{}
	\subfigure{\includegraphics[width=0.5\textwidth]{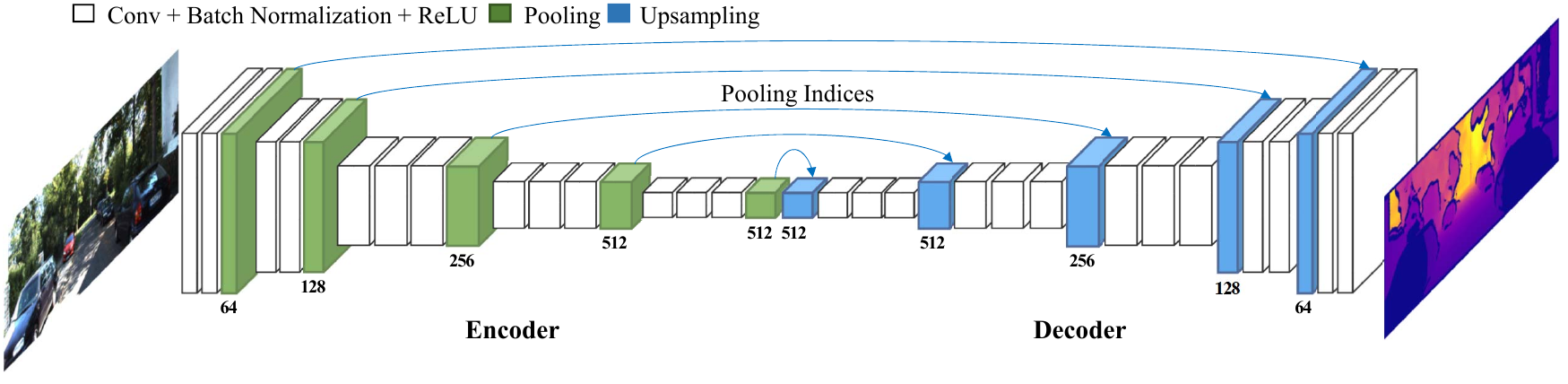}}\\
	\vspace{-3pt}
	\caption{Monocular depth estimation network.
		We designed a variant of the U-Net architecture~\cite{Unet} as a baseline network.}\vspace{-8pt}
	\label{fig:5}
\end{figure}

The pseudo-ground-truth depth maps inevitably contain erroneous estimates. 
To prevent such inaccurate depth values from being used when training the monocular depth estimation network, it is necessary to identify them in the pseudo-ground-truth depth maps. 
To this end, we additionally estimate the stereo confidence map~\cite{poggi16}. 
The confidence measure network extracts a square patch from the depth map and forwards it to a CNN to infer a normalized confidence value $C(p)$ $\in$ [0, 1].
We denote the confidence threshold as the hyper-parameter $\tau$.
The depth value of each pixel is set to be reliable when $C(p)\geq\tau$, and vice versa.
By adjusting $\tau$, we can control the sparseness and reliability of the depth map.
As $\tau$ increases, unreliable areas such as texture-less regions and occlusions are removed effectively, however, the depth map becomes sparse.
Sample images of depth maps and a stereo confidence map are shown in Fig.~\ref{fig:4}.
Because the confidence measure network uses a square patch extracted from a depth map without using padding or stride, it assigns zero to the boundary of the confidence map.
{
	Note that a rich line of research that successfully improves the performance when imperfect training data are used has been published~\cite{ding2018semi,li2019learning,guo2018learning,tosi2019learning}. 
	They showed that the use of massive pseudo ground truth data achieved outstanding results.
	Although the use of massive training data does not deliver perfect results, the performance can be significantly boosted by identifying erroneous data well.
	Our method is conceptually similar to these approaches.

		\begin{figure*}
			\centering
			\renewcommand{\thesubfigure}{}
			\subfigure{\includegraphics[width=0.195\textwidth,height=0.06\textheight]{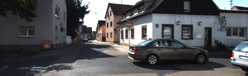}}
			\subfigure{\includegraphics[width=0.195\textwidth,height=0.06\textheight]{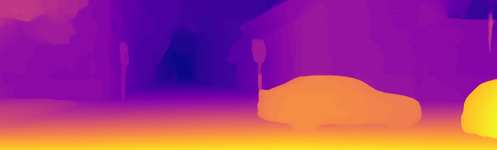}}
			\subfigure{\includegraphics[width=0.195\textwidth,height=0.06\textheight]{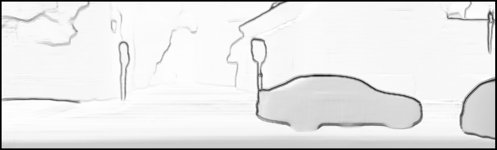}}
			\subfigure{\includegraphics[width=0.195\textwidth,height=0.06\textheight]{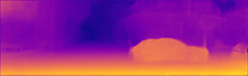}}
			\subfigure{\includegraphics[width=0.195\textwidth,height=0.06\textheight]{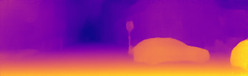}}
			\\ \vspace{-8pt}
			\subfigure[(a)]{\includegraphics[width=0.195\textwidth,height=0.06\textheight]{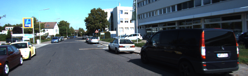}}
			\subfigure[(b)]{\includegraphics[width=0.195\textwidth,height=0.06\textheight]{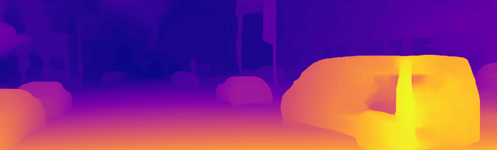}}
			\subfigure[(c)]{\includegraphics[width=0.195\textwidth,height=0.06\textheight]{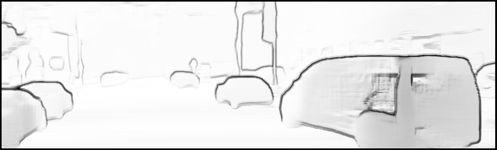}}
			\subfigure[(d)]{\includegraphics[width=0.195\textwidth,height=0.06\textheight]{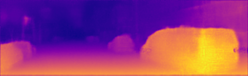}}
			\subfigure[(e)]{\includegraphics[width=0.195\textwidth,height=0.06\textheight]{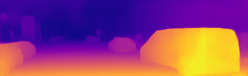}}
			\\ \vspace{-8pt}
			\caption{
				Visual results on the Eigen split~\cite{Eigen2015} of the KITTI dataset.
				(a) monocular left image, (b) pseudo-ground-truth depth maps, (c) estimated stereo confidence maps, (d) results of  depth maps trained without confidence guided loss, and (e) results of  depth maps trained with confidence guided loss.}
			\vspace{-13pt}
			\label{fig:6}			
		\end{figure*}

	\vspace{-4pt}	
	
	\subsection{Network Architecture}
	
	We designed monocular depth estimation networks based on a variant of the U-Net architecture~\cite{Unet}.
	As shown in Fig~\ref{fig:5}, the encoder network consists of the first 13 convolutional layers in the VGG~\cite{simonyan2014very} network, similar to~\cite{badrinarayanan2015segnet}.
	We discarded the fully connected layers in favor of maintaining spatial information.
	The convolutional layers consist of 3$\times$3 convolutions with batch normalization~\cite{Batchnorm} and a rectified linear unit (ReLU).
	Max-pooling with a 2$\times$2 window and a stride of 2 is performed at the encoder.
	The indices of the max locations are computed and stored during pooling. 
	Each convolutional layer at the encoder has a corresponding convolutional layer at the decoder; thus, the decoder network has 13 layers.
	The decoder upsamples the feature maps through unpooling with the memorized max-pooling indices~\cite{badrinarayanan2015segnet} from the corresponding encoder feature map.
	The sparse feature map is densified by convolving it with a trainable decoder filter bank.
	Using such pooling indices boosts the performance and enables more efficient training.
	The final decoder output is fed to a regression loss.

	\vspace{-4pt}
	
	\subsection{Training loss}
	
Given a monocular input image and pseudo-ground-truth depth map $\tilde{D}(p)$, we propose to use the \textit{stereo confidence guided regression loss} ${\cal L}_{c}$:

\begin{equation}\label{eq:1}
\centering{
	{{\cal L}_c} = \frac{1}{{\sum\limits_{p} M_{p} }}\sum\limits_{p} { M}_{p} \cdot {{{\left| {{{\hat D}(p)} - {{\tilde D}(p)}} \right|}_1}}} ,	
\end{equation}

\begin{equation}\label{eq:2}
\centering
{ M}_{p} = \left\{ {\begin{array}{*{20}{c}}
	{1,}&{{\rm{if }} \,C(p) \ge \tau }\\
	0,&{{\rm{if }} \,C(p) < \tau }
	\end{array}} \right.,
\end{equation}
where ${\hat{D}}(p)$ denotes the depth map predicted by the monocular depth estimation network.
In Section 4.4, we validate the effect of the stereo confidence measure by adjusting the hyper-parameter $\tau$.	
{
	The \textit{stereo confidence guided regression loss} in Eq.~\ref{eq:1} not only ensures the use of reliable depth values when training the monocular depth estimation network but also propagates reliable depth predictions from highly confident pixels into low-confidence pixels.
	This loss could handle the challenging elements of stereo matching such as occlusion, texture-less areas, and depth edges and effectively identifies reliable depth values.
	In Fig.~\ref{fig:6}, although there is no valid depth value around depth boundaries in terms of stereo confidence map, our results equipped with the stereo confidence guided loss show an excellent boundary preserving capability.
}

	\subsection{Large-scale outdoor stereo dataset} \label{3.5}
	
	Our new outdoor stereo dataset, the DIML/CVL RGB-D dataset, is complementary to existing RGB-D datasets such as the KITTI and Cityscapes datasets.
	To ensure the diversity of training data, we attempted to obtain non-driving scenes (\textit{e.g., }parks, buildings, apartments, trails, and streets) using hand-held stereo cameras, unlike the existing dataset, which consists mostly of driving scenes (\textit{e.g., }road and traffic scenes).
	Fig.~\ref{fig:7} shows sample RGB-D pairs of our dataset.
	The total number of distinct scenes is 1,053.
	Each scene contains a different number of images, ranging from 13,971 to 55,577.
	The images in this dataset were taken from fall 2015 to summer 2017 in four different cities (Seoul, Daejeon, Cheonan, and Sejong) in South Korea.
	
	Two types of stereo cameras, a ZED stereo camera~\footnote{\url{https://www.stereolabs.com/}} and a custom-built stereo camera, were used with different camera configurations for the baseline and focal length.
	The commercial ZED camera has a small baseline (12 cm), thus its sensing range is rather limited (up to 20 m).
	We designed a custom-built stereo system with mvBlueFox3 sensors~\footnote{\url{https://www.matrix-vision.com/USB3-vision-camera-mvbluefox3.html}} with a baseline of 40 cm to increase the maximum sensing range to 80 m.
	The stereo image was captured with a resolution of either 1920$\times$1080 or 1280$\times$720.
	Additional details of our dataset can be found in our technical report~\footnote{\url{https://dimlrgbd.github.io/downloads/technical\_report.pdf}}.
	All scenes were captured steadily with a tripod and slider in a hand-held fashion.

	The DIML/CVL RGB-D dataset comprises 1 million stereo images, depth maps computed using a stereo matching algorithm~\cite{MCCNN}, and a stereo confidence map~\cite{Skim2017,park2015leveraging}.
	The depth maps in the DIML/CVL RGB-D dataset were generated by MC-CNN~\cite{MCCNN}, which was a state-of-the-art stereo matching algorithm at the time of stereo image acquisition (from fall 2015 to summer 2017).
	Note that any type of stereo matching network can be adopted to obtain depth maps.
	The experiment we conducted using a more advanced stereo matching network is described in Section~\ref{4.3.3}.

	Our dataset differs from existing datasets in the following respects:
	\begin{enumerate}
		\item It is comprised of 1 million RGB-D data for outdoor scenes.
		\item Unlike existing outdoor datasets for driving scenes, ours was taken using hand-held stereo cameras for non-driving scenes.
		\item Stereo confidence maps are provided together to quantify the accuracy of depth maps. 
	\end{enumerate}

			\begin{figure}
				\centering
				\renewcommand{\thesubfigure}{}
				\subfigure{\includegraphics[width=0.117\textwidth,height=0.06\textheight]{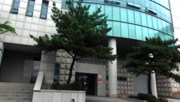}}
				\subfigure{\includegraphics[width=0.117\textwidth,height=0.06\textheight]{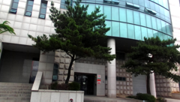}}
				\subfigure{\includegraphics[width=0.117\textwidth,height=0.06\textheight]{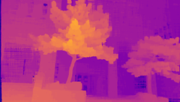}}
				\subfigure{\includegraphics[width=0.117\textwidth,height=0.06\textheight]{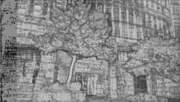}}
				\\ \vspace{-8pt}
				\subfigure{\includegraphics[width=0.117\textwidth,height=0.06\textheight]{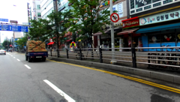}}
				\subfigure{\includegraphics[width=0.117\textwidth,height=0.06\textheight]{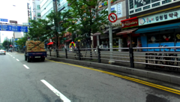}}
				\subfigure{\includegraphics[width=0.117\textwidth,height=0.06\textheight]{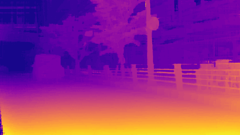}}
				\subfigure{\includegraphics[width=0.117\textwidth,height=0.06\textheight]{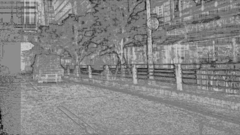}}
				\\ \vspace{-8pt}
				\subfigure[(a)]{\includegraphics[width=0.117\textwidth,height=0.06\textheight]{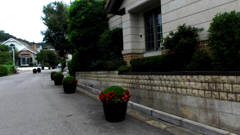}}
				\subfigure[(b)]{\includegraphics[width=0.117\textwidth,height=0.06\textheight]{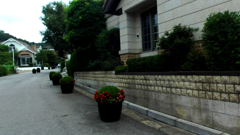}}
				\subfigure[(c)]{\includegraphics[width=0.117\textwidth,height=0.06\textheight]{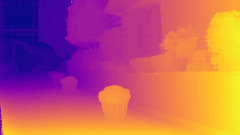}}
				\subfigure[(d)]{\includegraphics[width=0.117\textwidth,height=0.06\textheight]{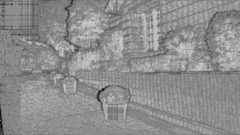}}
				\\
				\vspace{-8pt}
				\caption{Samples of RGB-D pairs with stereo confidence map from DIML/CVL dataset~\cite{DIML-CVL}. (a) left image, (b) right image, (c) depth map using~\cite{MCCNN}, and (d) stereo confidence map using~\cite{park2015leveraging}.}
				\label{fig:7}
				\vspace{-15pt}
			\end{figure}

	\subsection{Transfer of feature representation}
	
	In addition, we studied the effectiveness of our pre-trained monocular depth estimation network by transferring its feature representations as a pretext task for training other similar tasks such as road detection and semantic segmentation.
	The experimental results demonstrate that our pre-trained model is comparable to the ImageNet pre-trained model, which often serves as the pretext task for various vision applications~\cite{semi17,badrinarayanan2015segnet,MultiNet}.
	
	We fine-tuned our pre-trained model using the KITTI road benchmark~\cite{KITTI} for road detection and Cityscapes~\cite{Cityscape} for semantic segmentation, respectively.
	Both datasets include a small amount of manually annotated training data.
	We transfer both encoder and decoder weights of the pre-trained model into road detection and semantic segmentation.
	The softmax loss is used to fine-tune the network.

\begin{table*}
	\centering
	\caption{
		Quantitative evaluation of monocular depth estimation on the Eigen split~\cite{Eigen2015} of KITTI~\cite{KITTI} dataset.
		For dataset, K = KITTI, CS = Cityscapes, FT = FlyingThings, and DC = DIML/CVL.}
	\resizebox{0.99\textwidth}{!}{
		{\begin{tabular}{ccccccccccc}
				\toprule
				Method & Training data& Approach         & Dataset       & RMSE(lin)    & RMSE(log)    & Abs rel & Sqr rel &  $\delta < 1.25 $ & $\delta < 1.25^2$  &  $\delta < 1.25^3 $    \\  \cline{1-11}  \midrule
				& &                  &               & \multicolumn{4}{c}{Lower is better} & \multicolumn{3}{c}{Higher is better}    \\  \cline{1-11}
				&  & & & cap 80m  & & & & & \\
				\midrule
				Eigen \emph{et al.}~\cite{Eigen2015} & Left + LiDAR &  \textit{Sup.}     & K             & $7.156$ & $0.270$ & $0.215$ & $1.515$  & $0.692$           & $0.899$            & $0.967$  \\
				Godard \emph{et al.}~\cite{monodepth17} & Stereo &  \textit{Self-sup.}      & K             & $5.927$ & $0.247$ & $0.148$ & $1.344$  & $0.803$           & $0.922$            & $0.964$  \\
				Godard \emph{et al.}~\cite{monodepth17}& Stereo &  \textit{Self-sup.}    & K + CS            & $5.311$ & $0.219$ & $0.124$ & $1.076$  & $0.847$           & $0.942$            & $0.973$  \\
				Kuznietsov \emph{et al.}~\cite{semi17}& Left + LiDAR    & \textit{Sup.}      & K             & $4.815$ & $0.194$ & $0.122$ & $0.763$ &  $0.845$           & $0.957$            & \textbf{0.987} \\
				Kuznietsov \emph{et al.}~\cite{semi17}& Stereo + LiDAR  & \textit{Semi-sup}& K             & $4.621$ & $0.189$ & $0.113$ & $0.741$ &  $0.862$           & $0.960$            & $0.986$ \\
				Luo \emph{et al.}~\cite{SVS18}& (Synthetic) Stereo +  GT    &  \textit{Semi-sup.}& K + FT            & 4.681 & 0.200 & 0.102 & 0.700 &  0.872         & 0.954          & 0.978 \\
				Monodepth2~\cite{godard2019digging}& Stereo    & \textit{Self-sup.}  &    K    & 4.750  & 0.196 & 0.109 & 0.873 &   0.874        &    0.957       &  0.979 \\
				{Guo \emph{et al.}~\cite{guo2018learning}}& Left + Pseudo GT    & \textit{Semi-sup.}  &    K    & 4.634  & 0.189 & 0.105 & 0.811 &   0.874        &    0.959       &  0.982 \\
				{Tosi} \emph{et al.}~\cite{tosi2019learning}& Left + Pseudo GT    & \textit{Semi-sup.}  &    K    & 4.714  & 0.199 & 0.111 & 0.867 &   0.864        &    0.954       &  0.979 \\
				{Tosi} \emph{et al.}~\cite{tosi2019learning}& Left + Pseudo GT    & \textit{Semi-sup.}  &    K  + CS  & 4.351  & 0.184 & 0.096 & 0.673 &   0.890        &    0.961       &  0.981 \\
				\midrule
				Our Method& Left + Pseudo GT      & \textit{Semi-sup} & K            & 4.599           & 0.183          & 0.099  & 0.748 &   0.880         &     0.959      & 0.983  \\
				Our Method& Left + Pseudo GT     & \textit{Semi-sup} & K + DC      &  4.333     & 0.181     & 0.098  & 0.644 &  0.881       &     0.963      &  0.984 \\
				Our Method& Left + Pseudo GT     & \textit{Semi-sup} & K + CS    &  4.286     &  0.177   & 0.097  & 0.641  &  0.882       &     0.963      &  0.984 \\
				Our Method& Left + Pseudo GT     & \textit{Semi-sup} & K + CS + DC    &   \textbf{4.129}    &     \textbf{0.175}     & \textbf{0.095}  & \textbf{0.613}  & \textbf{0.884}       &  \textbf{0.964}    & 0.986 \\
				
				\midrule
				Upper Bound&       &  & K            & 3.475           & 0.158          & 0.055  & 0.418 &   0.941         &     0.969      & 0.982  \\
				\midrule
				&  & & & cap 50m  & & & & & \\
				\midrule
				Garg \emph{et al.}~\cite{garg2016unsupervised} & Stereo &  \textit{Self-sup.}     & K             & $5.104$ & $0.273$ & $0.169$ & $1.080$  & $0.740$           & $0.904$            & $0.962$  \\
				Godard \emph{et al.}~\cite{monodepth17}& Stereo &  \textit{Self-sup.}   & K             & $4.471$ & $0.232$ & $0.140$ & $0.976$  & $0.818$           & $0.931$            & $0.969$  \\
				Godard \emph{et al.}~\cite{monodepth17}& Stereo &  \textit{Self-sup.}    & K + CS            & $3.972$ & $0.206$ & $0.117$ & $0.762$  & $0.860$           & $0.948$            & $0.976$  \\
				Kuznietsov \emph{et al.}~\cite{semi17}& Stereo + LiDAR  & \textit{Semi-sup.} & K             & $3.518$ & $0.179$ & $0.108$ & $0.595$ &  $0.875$           & $0.964$            & \textbf{0.988} \\
				Luo \emph{et al.}~\cite{SVS18}&  (Sythetic) Stereo + GT    &  \textit{\textit{Semi-sup}}& K + FT            & 3.503 & 0.187 & 0.097 & 0.539 &  0.885         & 0.960          & 0.981 \\
				Monodepth2~\cite{godard2019digging}& Stereo    & \textit{Self-sup.}  &    K       & 3.868 & 0.127 & 0.092  & 0.537 &  0.892         &   0.962        &  0.986 \\
				Our Method& Left + Pseudo GT     & \textit{Semi-sup} & K + CS + DC    & \textbf{3.162}      & \textbf{0.162} & \textbf{0.091}   & \textbf{0.505}  & \textbf{0.901}          & \textbf{0.969}         & 0.986 \\
				\hline
				\bottomrule
			\end{tabular}}}
			\label{tab:1}
		\end{table*}

			\begin{figure*}
				\centering
				\renewcommand{\thesubfigure}{}
				\subfigure{\includegraphics[width=0.24\textwidth,height=0.045\textheight]{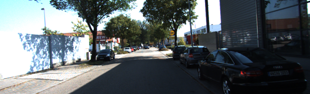}}
				\subfigure{\includegraphics[width=0.24\textwidth,height=0.045\textheight]{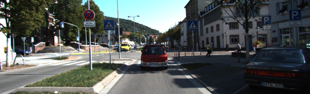}}
				\subfigure{\includegraphics[width=0.24\textwidth,height=0.045\textheight]{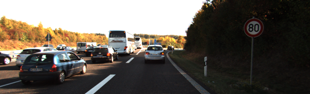}}
				\subfigure{\includegraphics[width=0.24\textwidth,height=0.045\textheight]{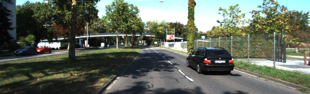}}
				\\  \vspace{-8pt}
				\subfigure{\includegraphics[width=0.24\textwidth,height=0.045\textheight]{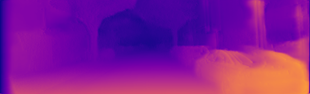}}
				\subfigure{\includegraphics[width=0.24\textwidth,height=0.045\textheight]{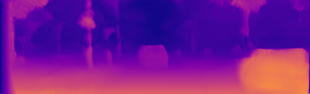}}
				\subfigure{\includegraphics[width=0.24\textwidth,height=0.045\textheight]{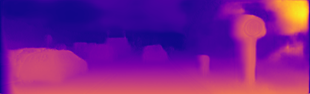}}
				\subfigure{\includegraphics[width=0.24\textwidth,height=0.045\textheight]{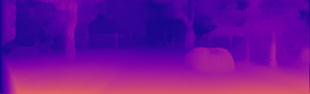}}
				\\  \vspace{-8pt}
				\subfigure{\includegraphics[width=0.24\textwidth,height=0.045\textheight]{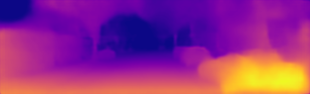}}
				\subfigure{\includegraphics[width=0.24\textwidth,height=0.045\textheight]{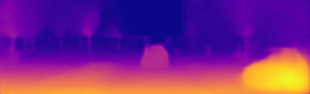}}
				\subfigure{\includegraphics[width=0.24\textwidth,height=0.045\textheight]{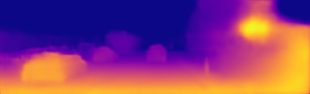}}
				\subfigure{\includegraphics[width=0.24\textwidth,height=0.045\textheight]{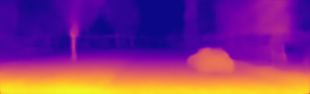}}
				\\      \vspace{-8pt}
				\subfigure{\includegraphics[width=0.24\textwidth,height=0.045\textheight]{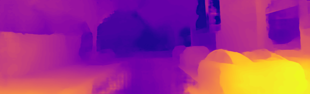}}
				\subfigure{\includegraphics[width=0.24\textwidth,height=0.045\textheight]{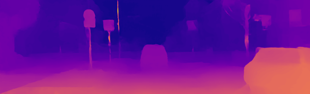}}
				\subfigure{\includegraphics[width=0.24\textwidth,height=0.045\textheight]{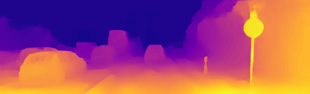}}
				\subfigure{\includegraphics[width=0.24\textwidth,height=0.045\textheight]{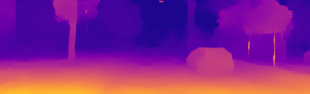}}
				\\      \vspace{-8pt}
				\subfigure{\includegraphics[width=0.24\textwidth,height=0.045\textheight]{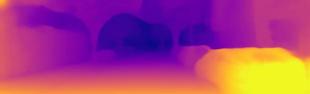}}
				\subfigure{\includegraphics[width=0.24\textwidth,height=0.045\textheight]{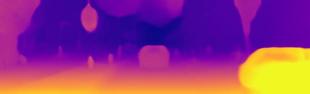}}
				\subfigure{\includegraphics[width=0.24\textwidth,height=0.045\textheight]{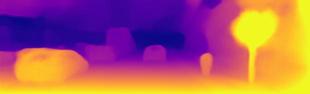}}
				\subfigure{\includegraphics[width=0.24\textwidth,height=0.045\textheight]{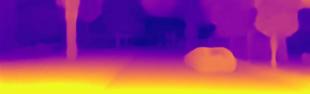}}
				\\  \vspace{-8pt}
				\subfigure{\includegraphics[width=0.24\textwidth,height=0.045\textheight]{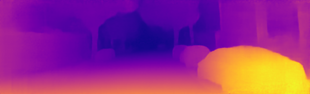}}
				\subfigure{\includegraphics[width=0.24\textwidth,height=0.045\textheight]{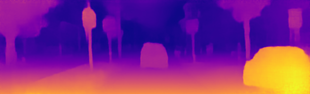}}
				\subfigure{\includegraphics[width=0.24\textwidth,height=0.045\textheight]{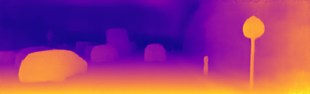}}
				\subfigure{\includegraphics[width=0.24\textwidth,height=0.045\textheight]{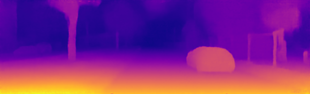}}
				\\      
				\vspace{-5pt}
				\caption{
					{Qualitative results on the Eigen split~\cite{Eigen2015} of the KITTI dataset~\cite{KITTI} (from top to bottom): input image, Godard \emph{et al.}~\cite{monodepth17}, Kuznietsov \emph{et al.}~\cite{semi17}, Luo \emph{et al.}~\cite{SVS18}, monodepth2~\cite{godard2019digging},and the proposed method.}}
				\label{fig:8}
				\vspace{-15pt}
			\end{figure*}

		\subsection{Implementation details}

		We adopted the stereo matching network (CRL)~\cite{CRL17} using the pre-trained model provided by the author and implemented the confidence measure network~\cite{poggi16}. 
		We implemented the monocular depth estimation network using the VLFeat MatConvNet library~\cite{Matconvnet}.
		
		\subsubsection{Confidence measure network}
		
		The network was trained using 50 image pairs consisting of the ground truth depth maps and stereo image pairs provided in the KITTI 2012 dataset.
		Following recent work on stereo confidence estimation~\cite{poggi16}, we trained the stereo confidence network with a relatively small number of image pairs. 
		This is possible because the confidence estimation can be seen as a local inference procedure.
		It determines whether an estimated depth value at a reference pixel is reliable when a small patch centered at the reference pixel is given. 
		Specifically, the confidence measure network is trained with a set of small patches (\textit{i.e.,} 9 $\times$ 9) that are paired with the pseudo-ground-truth depth map and ground truth confidence value~\cite{poggi16}.		
		The ground truth stereo confidence map is obtained by comparing the absolute difference between the predicted depth map and ground truth depth map (KITTI LiDAR points). 
		Following the literature~\cite{KITTI}, we set the confidence value to 1 when the absolute difference is smaller than 3 pixels, and 0 otherwise.
		
		During the training phase, we use a binary cross-entropy loss after applying a sigmoid function to the output of the network.
		We carried out 100 training epochs with an initial learning rate of 0.001, decreased by a factor {of} 10 every 10 epochs, and a momentum of 0.9.
		
		\subsubsection{Monocular depth network}
		To train the monocular depth estimation network, we collected stereo images from the KITTI, Cityscapes, and DIML/ CVL datasets, respectively.
		The pseudo-ground-truth training data were generated using the stereo matching network and stereo confidence map.
		The monocular student network was trained for 50 epochs with a batch size of 4.
		The Adam solver~\cite{Adam} was adopted for efficient stochastic optimization with a fixed learning rate of 0.001 and a momentum of 0.9.

	\section{Experimental Results}
	
	We conducted to validate the effectiveness of our semi-supervised monocular depth estimation via quantitative and qualitative comparisons with state-of-the-art methods in outdoor scenes. 
	For quantitative comparisons, we employ several metrics that have been used previously~\cite{Eigen2015,monodepth17,semi17,godard2019digging}:
	\vspace{-3pt}
	\begin{itemize}
		\renewcommand\labelitemi{\tiny$\bullet$}
		
		\item Threshold: $\%$ s.t. $\max \left( {\frac{{{d_i}}}{{{u_i}}},\frac{{{u_i}}}{{{d_i}}}} \right) = \delta  < thr$
		\item Abs rel:  $\frac{1}{N}\sum\nolimits_i {\left| {{d_i} - {u_i}} \right|/{d_i}}$
		\item Sqr rel: $\frac{1}{N}\sum\nolimits_i {{{\left\| {{d_i} - {u_i}} \right\|}^2}/{d_i}}$
		\item RMSE(lin): $\sqrt {\frac{1}{N}\sum\nolimits_i {{{\left\| {{d_i} - {u_i}} \right\|}^2}}}$
		\item RMSE(log): $\sqrt {\frac{1}{N}\sum\nolimits_i {{{\left\| {\log {d_i} - \log {u_i}} \right\|}^2}}}$
		\vspace{-3pt}			
	\end{itemize}
	where $u_i$ denotes the predicted depth at pixel $i$, and $N$ is the total number of pixels.
	The monocular depth network was trained by selecting outdoor scenes similar to recently reported state-of-the-art methods ~\cite{monodepth17,semi17,SVS18,godard2019digging}. 
	Indoor datasets were not used in our experiments, however, training with indoor data can be accomplished in the same manner.
	{
		In our method, the stereo matching (teacher) networks are trained with the small amount of labeled training data. 
		Then, the massive unlabeled training data, i.e., stereo image pairs, are used for training the monocular depth estimation (student) network. 
		Therefore, the proposed method can be categorized as a semi-supervised learning approach.}

\begin{figure*}
	\centering
	\renewcommand{\thesubfigure}{}
	\subfigure{\includegraphics[width=0.24\textwidth]{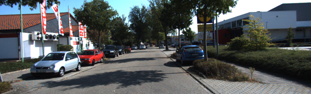}}
	\subfigure{\includegraphics[width=0.24\textwidth]{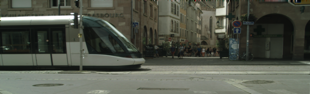}}
	\subfigure{\includegraphics[width=0.24\textwidth]{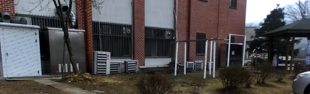}}
	\subfigure{\includegraphics[width=0.24\textwidth]{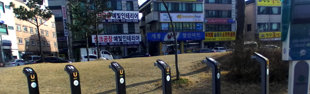}}
	\\ \vspace{-8pt}
	\subfigure{\includegraphics[width=0.24\textwidth]{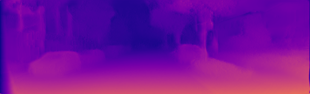}}
	\subfigure{\includegraphics[width=0.24\textwidth]{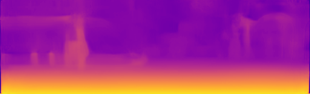}}
	\subfigure{\includegraphics[width=0.24\textwidth]{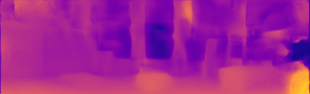}}
	\subfigure{\includegraphics[width=0.24\textwidth]{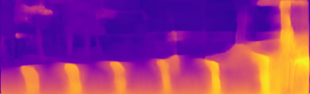}}
	\\ \vspace{-8pt}
	\subfigure{\includegraphics[width=0.24\textwidth]{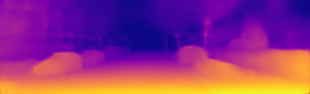}}
	\subfigure{\includegraphics[width=0.24\textwidth]{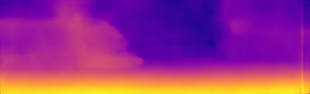}}
	\subfigure{\includegraphics[width=0.24\textwidth]{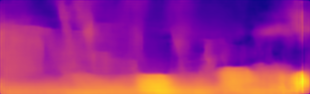}}
	\subfigure{\includegraphics[width=0.24\textwidth]{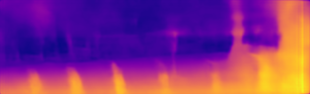}}
	\\ \vspace{-8pt}
	\subfigure{\includegraphics[width=0.24\textwidth]{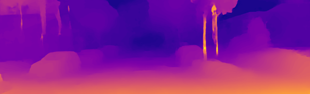}}
	\subfigure{\includegraphics[width=0.24\textwidth]{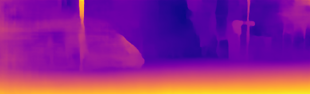}}
	\subfigure{\includegraphics[width=0.24\textwidth]{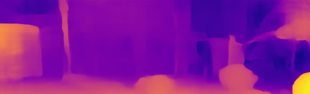}}
	\subfigure{\includegraphics[width=0.24\textwidth]{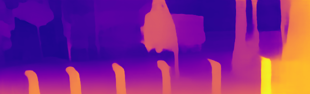}}
	\\ \vspace{-8pt}
	\subfigure{\includegraphics[width=0.24\textwidth]{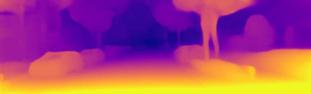}}
	\subfigure{\includegraphics[width=0.24\textwidth]{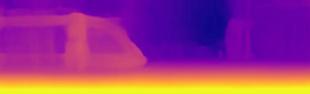}}
	\subfigure{\includegraphics[width=0.24\textwidth]{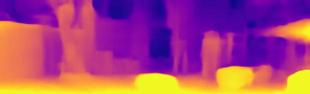}}
	\subfigure{\includegraphics[width=0.24\textwidth]{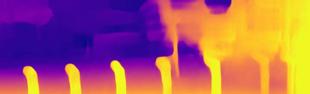}}
	\\ \vspace{-8pt}
	\subfigure{\includegraphics[width=0.24\textwidth]{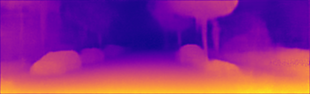}}
	\subfigure{\includegraphics[width=0.24\textwidth]{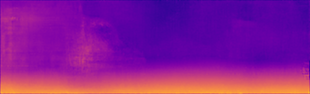}}
	\subfigure{\includegraphics[width=0.24\textwidth]{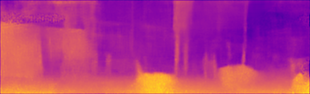}}
	\subfigure{\includegraphics[width=0.24\textwidth]{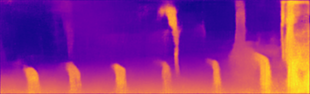}}
	\\ \vspace{-8pt}
	\subfigure{\includegraphics[width=0.24\textwidth]{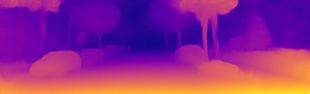}}
	\subfigure{\includegraphics[width=0.24\textwidth]{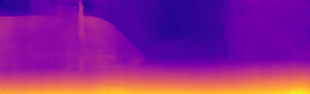}}
	\subfigure{\includegraphics[width=0.24\textwidth]{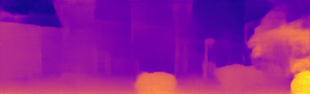}}
	\subfigure{\includegraphics[width=0.24\textwidth]{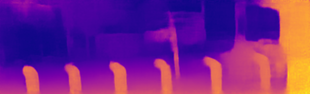}}
	\\ \vspace{-8pt}
	\subfigure[KITTI (Target)]{\includegraphics[width=0.24\textwidth]{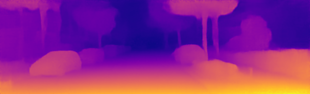}}
	\subfigure[Cityscape (Novel)]{\includegraphics[width=0.24\textwidth]{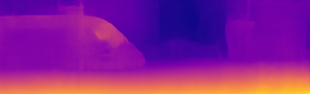}}
	\subfigure[DIML/CVL (Novel)]{\includegraphics[width=0.24\textwidth]{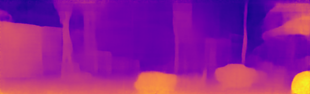}}
	\subfigure[DIML/CVL (Novel)]{\includegraphics[width=0.24\textwidth]{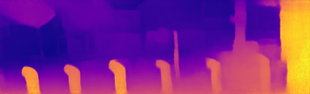}}
	\\			\vspace{-5pt}
	\caption{
		{		       		
			Qualitative comparison for the impact of scene diversity (from top to bottom): input image, Godard \emph{et al.}~\cite{monodepth17}, Kuznietsov \emph{et al.}~\cite{semi17}, Luo \emph{et al.}~\cite{SVS18}, monodepth2~\cite{godard2019digging}, and the proposed method trained with K, K + DC, KITTI + CS + DC, respectively.}}
	\label{fig:9}
	\vspace{-15pt}
\end{figure*}

	\subsection{Dataset}

	We generated pseudo-ground-truth depth maps using stereo images provided in the KITTI~\cite{KITTI}, Cityscapes~\cite{Cityscape}, and DIML/CVL datasets.
	All images were resized to 620$\times$188 for training and testing.

		\subsubsection{KITTI (K)}
		
		This dataset consists of outdoor driving scenes with sparse depth maps captured by the Velodyne LiDAR.
		The depth map is very sparse (density of less than 6\%), and depth values are available only at the bottom parts of a color image.
		The dataset contains 42,382 rectified stereo pairs from 61 scenes, with a typical image sized 1242$\times$375 pixels.
		Following the Eigen split~\cite{Eigen2015}, we split the stereo image pairs into 22,600 images for training, 888 images for validation, and 697 images for testing.

		\subsubsection{Cityscapes (CS)}
		
		This dataset was originally constructed for semantic segmentation and provides manually annotated segmentation maps for 19 semantic classes, consisting of 2,975 images for training, 500 images for validation, and 1,525 images for testing.
		Additionally, they provide 22,973 stereo image pairs with spatial dimensions of 2048$\times$1024.
		We split the stereo image pairs into 21,283 for training, 500 for validation, and 3,215 for testing.
		Following the literature~\cite{monodepth17}, we cropped the stereo images by discarding the bottom  20\% (the car hood) and then resized them.

		\subsubsection{DIML/CVL (DC)}
		
		Our outdoor dataset consists of 1 million stereo image pairs, depth maps, and stereo confidence maps.
		The original spatial resolution of this dataset is 1920$\times$1080 or 1280$\times$720.
		Following previous practices~\cite{Cityscape,KITTI}, of 1 million image pairs, we selected 23,500 image pairs.
		We split them into 22,000 images for training, 800 images for validation, and 700 images for testing.
		Input images were cropped and resized by discarding the bottom parts containing mostly ground.

		\subsection{Comparison with state-of-the-art methods}

		We compared existing monocular depth estimation approaches including supervised~\cite{Eigen2015,semi17}, self-supervised~\cite{monodepth17,godard2019digging}, and semi-supervised methods~\cite{semi17,SVS18} with the proposed approach.
		{Of the 1 million stereo image pairs, we selected 22,000 stereo images from various scenes as training data, similar to Eigen split~\cite{Eigen2015}.}
		The results in Table \ref{tab:1} indicate that our method consistently outperforms recent approaches, except for $\delta < 1.25^3$.
		Following the literature~\cite{monodepth17,semi17,SVS18,godard2019digging}, the depth value was truncated at 80m or 50m.
		The proposed method was trained using various combinations of KITTI, Cityscapes, and our dataset.
		Eigen \emph{et al.}~\cite{Eigen2015} was trained using ground-truth depth maps augmented from the KITTI 2015 dataset.
		Godard \emph{et al.}~\cite{monodepth17} and monodepth2~\cite{godard2019digging} proposed a self-supervised approach in which stereo images of the KITTI and/or Cityscapes dataset were used.
		Although this method requires no ground truth depth maps during training, it is difficult to handle occlusions effectively and does not obtain a sharp depth boundary owing to the limitation of the image reconstruction loss.
		Kuznietsov \emph{et al.}~\cite{semi17} employed both supervised regression loss and unsupervised reconstruction loss~\cite{monodepth17} to achieve a performance gain over existing supervised and self-supervised approaches~\cite{Eigen2015, monodepth17}.
		However, their method still requires ground truth depth maps as supervision.
		Luo \emph{et al.}~\cite{SVS18} proposed the use of a view synthesis network and stereo matching network in a unified framework.
		However, their model was mainly trained with synthetic FlyingThings3D data~\cite{DispNet}, and thus incurs domain gaps between synthetic and realistic images and cannot generalize well on real data.
		Such two-step method substantially increases the computational complexity (0.53 sec for an inference).			
		
			Guo \emph{et al.}~\cite{guo2018learning} trained the stereo matching network (teacher) using a synthetic stereo dataset.
			The pseudo labels for monocular depth estimation may thus incur undesired artifacts due to a large domain gap between synthetic and real datasets when training the monocular depth estimation network. 
			To alleviate this domain gap, they performed the unsupervised fine-tuning.
			However, they do not fully consider the inherent errors such as occlusion, texture-less region, and repetitive patterns that occur in the stereo matching. 
			Therefore, the performance of the generated pseudo label is limited.
			The method in Tosi \emph{et al.}~\cite{tosi2019learning} is somewhat similar to the proposed method in that pseudo labels are refined by removing matching outliers through a left-right consistency check. 
		{
				However, the simple consistency check often fails to exclude the matching outliers of the pseudo depth labels which are constructed by a handcrafted stereo matching algorithm. 
				In contrast, our method attempts to leverage such stereo knowledge more effectively with the proposed confidence guided loss and data ensemble strategy. 
				The confidence guided loss can resolve the performance degeneration incurred by unreliable pseudo depth values in occlusion, texture-less region, and repetitive pattern that cannot be addressed using stereo matching methods only. 
				The data ensemble method also helps improve the quality of the pseudo labels by combining multiple depth predictions beyond a single scale prediction. 
				Furthermore, Tosi et al..~\cite{tosi2019learning} leverages a correlation layer to generate 3D cost volume from the input left image and synthesized right image during an inference, which demands a high computational complexity (0.48 sec for an inference). 
				In contrast, our model can retain a relatively low complexity (0.21 sec) as the stereo knowledge is imposed only on the label (depth) space by training the student network with pseudo depth maps already computed from the teacher network.
			}
		
		Furthermore, qualitative results are provided in Fig.~\ref{fig:8}.
		Our monocular depth network achieves more accurate and edge-preserved depth maps than state-of-the-art methods ~\cite{monodepth17, semi17, SVS18,godard2019digging}.

\begin{table}
	\centering
	\caption{{Quantitative evaluation of monocular depth estimation trained with the amount of data differently.
			The number in the bracket indicates the amount of training data.}}
	\scalebox{0.83}{
		\begin{tabular}{ccccc}
			\toprule
			{Method}  & {Training data}  & {RMSE(lin)} & {RMSE(log)} & { Abs rel }  \\ \hline
			Our Method     & CS (20k)     &  6.328  & 0.284  &   0.206         \\
			Our Method    & DC (20k)  &  7.944 & 0.339  &  0.361           \\
			Our Method    & DC (100k)  & 5.647  & 0.236 &  0.161           \\
			Our Method    & DC (200k)  &  5.250 & 0.217  &  0.143           \\
			Our Method     & K + DC (200k)  & 4.194 & 0.176 &  0.096          \\
			\bottomrule
		\end{tabular}}
		\label{tab:2}
		\vspace{-8pt}
	\end{table}

		\subsection{Discussion}
		
		\subsubsection{Impact of scene diversity of DIML/CVL dataset}
		
		We conducted an ablation study to demonstrate that the DIML/CVL dataset is able to effectively train the algorithm in our semi-supervised approach.
		Fig.~\ref{fig:9} shows the qualitative results obtained by the proposed method for two different combinations (\nth{6} to \nth{8} rows) of pseudo-ground-truth training data. 
		Similar results were obtained in both the source domain and a novel domain when using the DIML/CVL dataset for training.
		This indicates that the proposed method addresses the generalization issue well when used in a novel domain. 
		Specifically, using the DIML/CVL dataset for training leads to a performance gain in the novel domain, for example, when comparing the results of the proposed method trained with KITTI (\nth{6} row) and KITTI+DIML/CVL (\nth{7} row). 
		The proposed method consistently generates smooth depth maps with sharp edges and recovers the overall scene layout well. 
		In addition, we included the results (\nth{2} to \nth{5} rows) of state-of-the-art approaches~\cite{SVS18,monodepth17,semi17,godard2019digging} for a qualitative comparison. 
		{When inferring with the Cityscapes and DIML/CVL datasets (novel domain), they exhibited significant decreases in the performance.
			It was possible to train two of the methods~\cite{monodepth17,godard2019digging} with the Cityscapes and DIML/CVL dataset providing stereo image pairs, as these methods do not require ground truth depth maps for training. 
			In contrast, the other two methods we included in our evaluation~\cite{SVS18} and~\cite{semi17} could not be trained with these datasets. 
			The self-supervised approaches~\cite{monodepth17,godard2019digging} can be learned using training data from both the source and novel domains, but their self-supervised loss incurs blurry depth boundaries and inaccurate estimates in the occlusion, as revealed in the \nth{2} row and \nth{5} row in Fig.~\ref{fig:9}.}

		\begin{figure*}
			\centering
			\renewcommand{\thesubfigure}{}
			\subfigure{\includegraphics[width=0.13\textwidth,height=0.12\textheight]{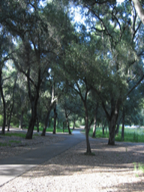}}
			\subfigure{\includegraphics[width=0.13\textwidth,height=0.12\textheight]{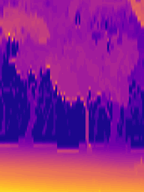}}
			\subfigure{\includegraphics[width=0.13\textwidth,height=0.12\textheight]{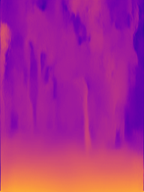}}
			\subfigure{\includegraphics[width=0.13\textwidth,height=0.12\textheight]{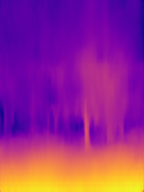}}
			\subfigure{\includegraphics[width=0.13\textwidth,height=0.12\textheight]{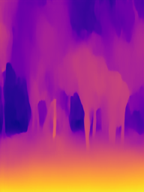}}
			\subfigure{\includegraphics[width=0.13\textwidth,height=0.12\textheight]{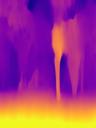}}
			\subfigure{\includegraphics[width=0.13\textwidth,height=0.12\textheight]{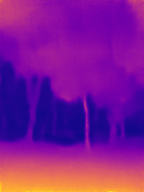}}
			\\				\vspace{-8pt}
			\subfigure[(a)]{\includegraphics[width=0.13\textwidth,height=0.12\textheight]{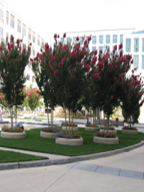}}
			\subfigure[(b)]{\includegraphics[width=0.13\textwidth,height=0.12\textheight]{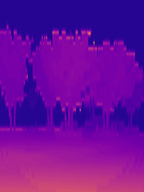}}
			\subfigure[(c)]{\includegraphics[width=0.13\textwidth,height=0.12\textheight]{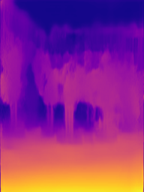}}
			\subfigure[(d)]{\includegraphics[width=0.13\textwidth,height=0.12\textheight]{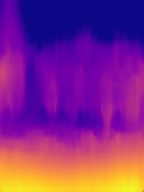}}
			\subfigure[(e)]{\includegraphics[width=0.13\textwidth,height=0.12\textheight]{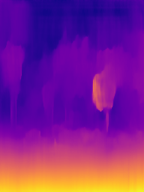}}
			\subfigure[(f)]{\includegraphics[width=0.13\textwidth,height=0.12\textheight]{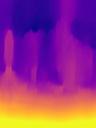}}
			\subfigure[(g)]{\includegraphics[width=0.13\textwidth,height=0.12\textheight]{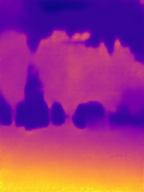}}	
			\\ \vspace{-5pt}
			\caption{Qualitative results on the Make3D dataset~\cite{Make3D}: (a) input image, (b) ground truth depth maps, (c) Godard \emph{et al.}~\cite{monodepth17}, (d) Kuznietsov \emph{et al.}~\cite{semi17}, (e) Luo \emph{et al.}~\cite{SVS18}, (f) monodepth2~\cite{godard2019digging}, and (g) the proposed method. 
				None of the models were trained on Make3D.}
			\label{fig:10}
			\vspace{-8pt}
		\end{figure*}

		Table~\ref{tab:2} reports the depth accuracy using the Eigen split in the KITTI dataset. 
		We conducted an experiment to verify the effectiveness of the DIML/CVL dataset according to the amount of training data.
		Because ground truth depth maps are provided in the KITTI dataset, the objective evaluation was conducted using this dataset.
		First, we trained our monocular depth network with the Cityscape and DIML/CVL datasets. 
		With the same amount of data, the model trained with the Cityscape dataset is more effective than that trained with the DIML/CVL dataset. 
		This is because the Cityscape dataset mostly contains driving scenes that are relatively similar to those in KITTI, whereas the DIML/CVL dataset includes both driving and non-driving scenes. 
		Interestingly, when we increase the amount of DIML/CVL training data, the depth accuracy gradually improves even though the majority of images in the DIML/CVL dataset consist of non-driving scenes. 
		This demonstrates the complementarity of the DIML/CVL data. 
		Although we trained the monocular depth network using up to 200,000 images for training, the use of a larger amount of training data would also be possible.

		\begin{figure*}
			\centering
			\renewcommand{\thesubfigure}{}
			\subfigure{\includegraphics[width=0.152\textwidth,height=0.08\textheight]{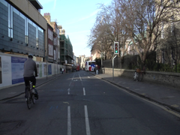}}
			\subfigure{\includegraphics[width=0.152\textwidth,height=0.08\textheight]{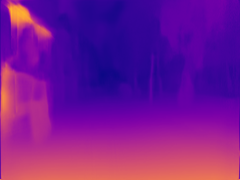}}
			\subfigure{\includegraphics[width=0.152\textwidth,height=0.08\textheight]{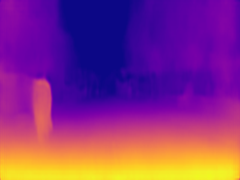}}
			\subfigure{\includegraphics[width=0.152\textwidth,height=0.08\textheight]{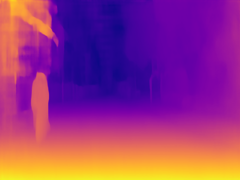}}
			\subfigure{\includegraphics[width=0.152\textwidth,height=0.08\textheight]{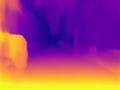}}
			\subfigure{\includegraphics[width=0.152\textwidth,height=0.08\textheight]{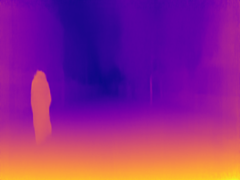}}
			\\			\vspace{-8pt}
			\subfigure[(a)]{\includegraphics[width=0.152\textwidth,height=0.08\textheight]{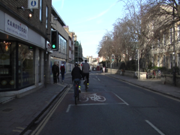}}
			\subfigure[(b)]{\includegraphics[width=0.152\textwidth,height=0.08\textheight]{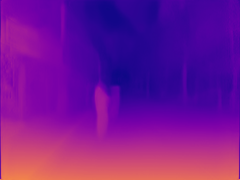}}
			\subfigure[(c)]{\includegraphics[width=0.152\textwidth,height=0.08\textheight]{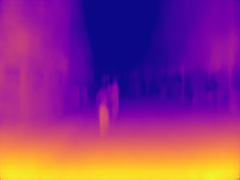}}
			\subfigure[(d)]{\includegraphics[width=0.152\textwidth,height=0.08\textheight]{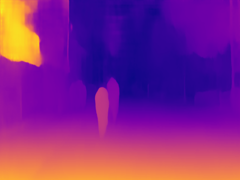}}
			\subfigure[(e)]{\includegraphics[width=0.152\textwidth,height=0.08\textheight]{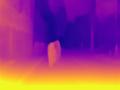}}
			\subfigure[(f)]{\includegraphics[width=0.152\textwidth,height=0.08\textheight]{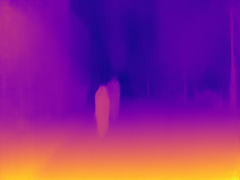}}
			\\ \vspace{-5pt}
				\caption{Qualitative results on the CamVid dataset~\cite{Camvid}: (a) input image, (b) Godard \emph{et al.}~\cite{monodepth17}, (c) Kuznietsov \emph{et al.}~\cite{semi17}, (d) Luo \emph{et al.}~\cite{SVS18}, (e) monodepth2~\cite{godard2019digging}, and (f) the proposed method. None of the models were trained on CamVid.}
				\label{fig:11}
				\vspace{-10pt}
		\end{figure*}

		\subsubsection{Generalization to other datasets}	
		
		{Following the existing literatures~\cite{godard2019digging,monodepth17,zhao2019geometry}, we used two datasets to validate the generalization ability of our method to other datasets, Make3D~\cite{Make3D} and camVid.}

		Similar to previous work~\cite{monodepth17}, we cropped the top and bottom of the input images in Make3D dataset for testing.
		As for the ground truth depth maps, we reshaped and cropped the top and bottom to match our predictions. 
		Owing to the limitation of the depth-sensing device that was used for capturing depth data, we removed depth values larger than 70 m. 
		Table~\ref{tab:3} lists the numerical results, including comparisons to several methods trained on non-Make3D training data. 
		Our approach achieves the best performance among all non-Make3D-trained models. 
		It should be noted that our results are more accurate than or comparable to those of the state-of-the-art methods even when trained with the KITTI dataset only. 
		This indicates the effectiveness of using pseudo-ground-truth depth maps and a stereo confidence map for training the monocular depth network. 
		Using DIML/CVL leads to a meaningful performance gain, showing the importance of our training data. 
		Fig.~\ref{fig:10} visualizes the results of the depth predictions. 
		Our results preserve the overall structure and boundary of objects more accurately than other depth maps, and the resulting images we obtain with our method are the most similar to the ground truth depth maps.
					
						\begin{table}
							\centering
							\caption{{Quantitative evaluation of monocular depth estimation on the Make3D test set for various approaches.}}
							\label{my-label}
							{    \scalebox{0.8}{
									\begin{tabular}{ccccc}
										\toprule
										{Method}  & {Dataset}  & {RMSE(lin)} & {RMSE(log)} & { Absrel }  \\ \hline
										Godard \emph{et al.}~\cite{monodepth17}  &  K  & 11.762  & 0.193     &   0.544        \\
										Godard \emph{et al.}~\cite{monodepth17}   &  K + CS  &  10.369 & 0.188     &    0.536       \\
										Kuznietsov \emph{et al.}~\cite{semi17}  &  K  & 8.237  & 0.191 &  0.421  \\
										Luo \emph{et al.}~\cite{SVS18}  & K + FT   & 8.184  & 0.185  &  0.428           \\
										monodepth2~\cite{godard2019digging}  &  K   &  8.238 & 0.201     &    0.374       \\
										\midrule
										Our Method     & K     & 7.992  & 0.181  &    0.423         \\
										Our Method    & DC  & 5.512  & 0.118 & 0.297            \\
										Our Method     & K + DC  &  5.347 & 0.113 & 0.278              \\
										Our Method     & K + CS + DC  & 5.136 & 0.107 & 0.265           \\
										\bottomrule
									\end{tabular}}}
									\label{tab:3}
								\end{table}

		Because the CamVid driving dataset~\cite{Camvid} only provides ground truth maps for semantic segmentation, we validate the effectiveness of our method by carrying out qualitative comparisons with state-of-the-art methods. 
		In Fig.~\ref{fig:11}, despite the domain gap between the training and test images in terms of location, image characteristics, and camera parameters, our model still produces visually plausible depths in a novel domain.

			\subsubsection{Impact of the stereo matching network} \label{4.3.3}

			Any stereo matching methods can be utilized as the teacher network in our framework.
			For more solid analysis, we conducted the impact of the stereo matching network using various stereo matching algorithms.
			{Table~\ref{tab:stereo} shows the quantitative performance of depth accuracy and runtime of the stereo matching network.}
			The accuracy of pseudo ground truth depth results from MC-CNN and CRL shows a large performance gap.
			This performance gap also brings a large gap in monocular depth accuracy (MC-CNN (RMSE (lin)): 5.294/ CRL (RMSE (lin)): 4.995).
			Although the pseudo depth maps of GA-Net has a higher accuracy than those of CRL, the performance gain in terms of the monocular depth accuracy was relatively marginal (GA-Net (RMSE (lin)): 4.946).
			{This is because the confidence map excludes unreliable pseudo depth values effectively. 
				Additionally, the fast inference time of CRL enables us to construct the pseudo RGB+D dataset efficiently.
				Thus, we adopted CRL as a teacher network in our framework.}

			\begin{figure}
				\renewcommand{\thesubfigure}{}
				\subfigure{\includegraphics[width=0.32\textheight,height=0.34\textwidth]{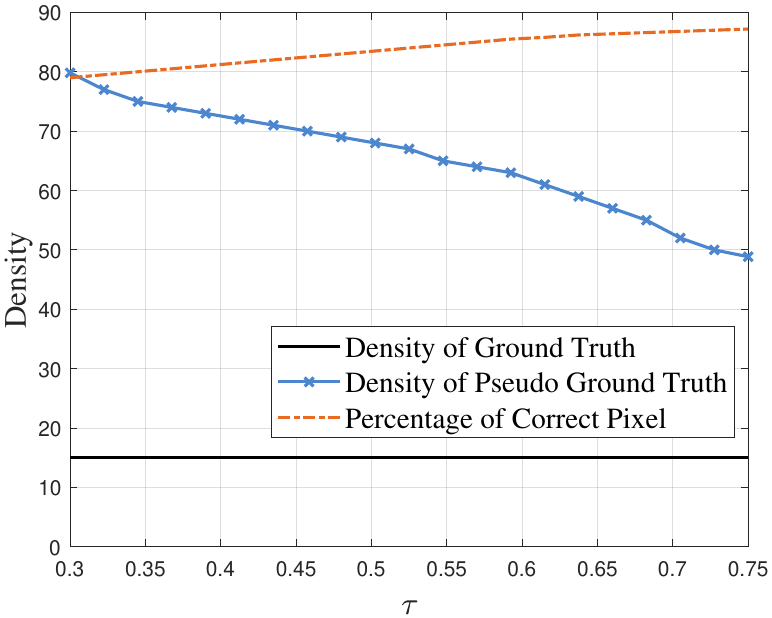}}
				\vspace{-8pt}
				\caption{{Quality analysis of pseudo-ground-truth depth maps controlled by the confidence threshold $\tau$.}}
				\label{fig:12}

			\end{figure}

		\begin{table}
			\centering
			\caption{{Quantitative evaluation of the impact of the stereo matching network.}}
			{    \scalebox{1}{
					\begin{tabular}{cccc}
						\toprule
						{Method}  & {RMSE(lin)}  & { Absrel }  & Time(s) \\ \hline
						MC-CNN~\cite{MCCNN}  & 4.236   & 0.074 & 67      \\
						CRL~\cite{CRL17}   &  3.578 & 0.055  &  0.47 \\
						GA-NET~\cite{GAnet}  &  3.457  & 0.050 & 1.5 \\
						\bottomrule
					\end{tabular}}}
					\label{tab:stereo}
					\vspace{-15pt}
				\end{table}

	\subsection{Ablation Study}

	\subsubsection{Impact of label quality}
	
	We first investigated the performance gain of the data ensemble and the tradeoff between accuracy and density that determines the performance of the pseudo-ground-truth depth maps. 
	The KITTI dataset using the Eigen split~\cite{Eigen2015} was used {for an objective evaluation.}
	The density and accuracy of the pseudo-ground-truth depth maps are controlled by the confidence threshold. 
	{However, the confidence map of the pseudo-ground-truth depth maps is not perfect.}
	Nevertheless, excluding the depth outliers with the confidence map greatly improves the performance of the monocular depth estimation. 
	This can be explained by two factors: 1) the quality of the pseudo-ground-truth depth maps in Fig.~\ref{fig:12} and 2) the accuracy of the monocular depth estimation in Fig.~\ref{fig:13}.
	
	In Fig.~\ref{fig:12}, we present the tradeoff between the accuracy and density of the pseudo-ground-truth depth maps.
	{The black horizontal line shows the density of the KITTI ground truth depth maps. 
		{When the estimated confidence value $C(p)\in [0, 1]$ at pixel $p$, we denote a set of pixels estimated to be correct ($C(p)$ $\ge$ $\tau$) in the pseudo-ground-truth depth map as $P$.}
		Note that the ground truth depth map is sparse, and thus, this is computed only at a set of valid depth pixels $G$ in the ground truth depth map. Subsequently, the density of the pseudo-ground-truth depth map is computed using $|P|/|G|$. The accuracy is computed using $|A|/|P|$, where $A$ indicates a set of correct pixels among $P$.}
	The larger the value of $\tau$, the higher the accuracy of the pseudo-ground-truth depth maps becomes. 
	However, this reduces the density of the depth maps. 
	For instance, when $\tau=0.75$, only half of the depth pixels are chosen as reliable.

	\begin{figure}[]
		\renewcommand{\thesubfigure}{}
		\subfigure{\includegraphics[width=0.35\textheight,height=0.32\textwidth]{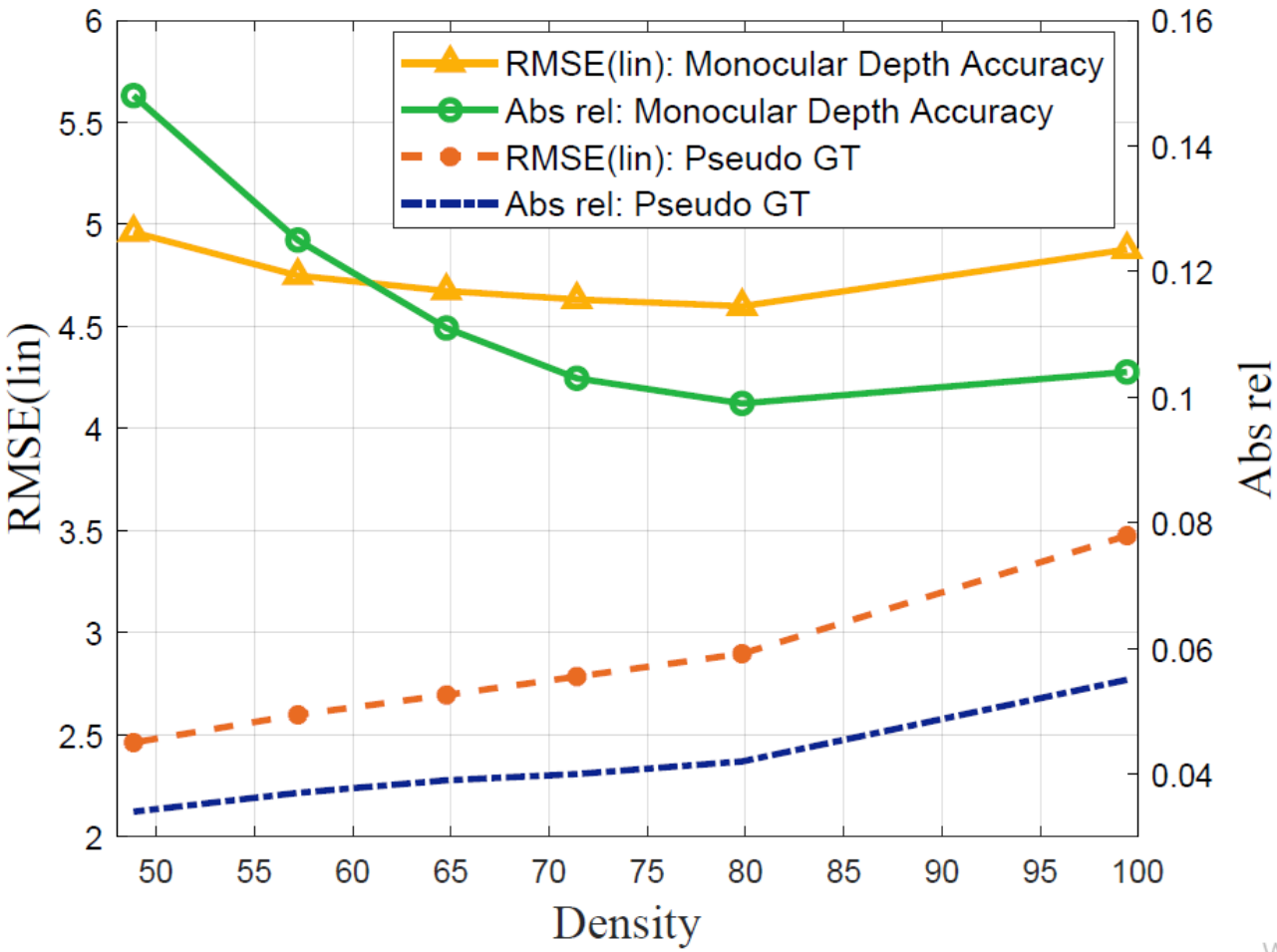}}
		\vspace{-8pt}
		\caption{
			{Analysis of the trade-off between the accuracy and density of pseudo-ground-truth depth maps.}}
		\label{fig:13}
		\vspace{-15pt}
	\end{figure}

					\begin{figure*}[!]
						\centering
						\renewcommand{\thesubfigure}{}
						\subfigure{\includegraphics[width=0.19\textwidth]{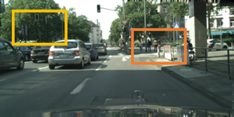}}
						\subfigure{\includegraphics[width=0.19\textwidth]{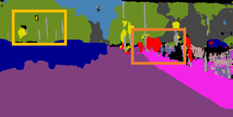}}
						\subfigure{\includegraphics[width=0.19\textwidth]{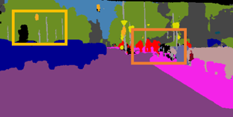}}
						\subfigure{\includegraphics[width=0.19\textwidth]{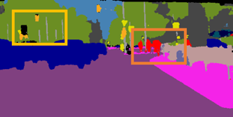}}
						\subfigure{\includegraphics[width=0.19\textwidth]{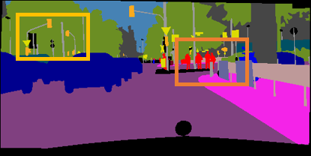}}\\
						\vspace{-8pt}
						\subfigure{\includegraphics[width=0.19\textwidth]{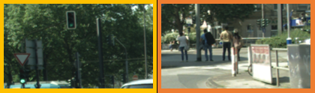}}
						\subfigure{\includegraphics[width=0.19\textwidth]{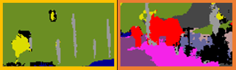}}
						\subfigure{\includegraphics[width=0.19\textwidth]{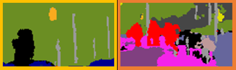}}
						\subfigure{\includegraphics[width=0.19\textwidth]{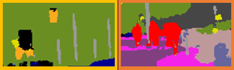}}
						\subfigure{\includegraphics[width=0.19\textwidth]{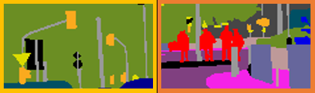}}\\
						\vspace{-8pt}
						\subfigure{\includegraphics[width=0.19\textwidth]{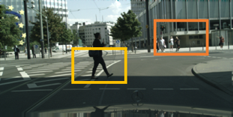}}
						\subfigure{\includegraphics[width=0.19\textwidth]{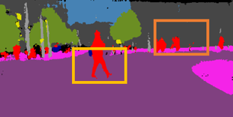}}
						\subfigure{\includegraphics[width=0.19\textwidth]{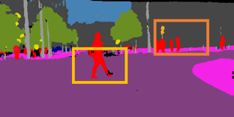}}
						\subfigure{\includegraphics[width=0.19\textwidth]{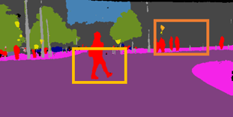}}
						\subfigure{\includegraphics[width=0.19\textwidth]{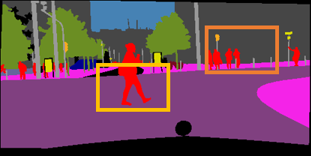}}\\
						\vspace{-8pt}
						\subfigure[(a)]{\includegraphics[width=0.19\textwidth]{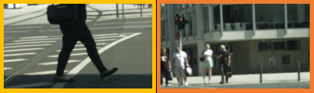}}
						\subfigure[(b)]{\includegraphics[width=0.19\textwidth]{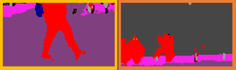}}
						\subfigure[(c)]{\includegraphics[width=0.19\textwidth]{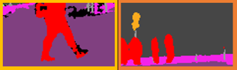}}
						\subfigure[(d)]{\includegraphics[width=0.19\textwidth]{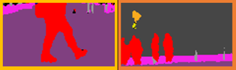}}
						\subfigure[(e)]{\includegraphics[width=0.19\textwidth]{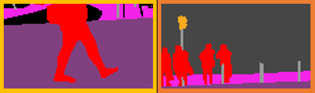}}\\ \vspace{-5pt}
						\caption{						
							Semantic segmentation results on the Cityscapes dataset: (a) input images, (b) $\sim$ (d) results of fine-tuning with different initialization methods. 
							(b) scratch, (c) ImageNet pre-trained model~\cite{simonyan2014very}, (d) our pre-trained model, and (e) ground truth image.
							The results indicate the benefit of using our pre-trained model.}
						\label{fig:14}
						\vspace{-15pt}
					\end{figure*}

		Fig.~\ref{fig:13} shows the accuracy measurements of the monocular depth network according to the confidence threshold $\tau$. 
		Fig.~\ref{fig:13} shows the RMSE(lin) and Abs rel of pseudo ground truth depth maps and monocular depth maps.
		{
			We use a stereo confidence map to identify inaccurate depth values (Fig.~\ref{fig:8}) and avoid these values when training the monocular depth estimation network. 
			Here, we investigate the inference accuracy of the monocular depth estimation according to the density of the pseudo-ground-truth depth maps. Training with more accurate pseudo-ground-truth depth maps (yet with a lower density) does not necessarily increase the accuracy of the monocular depth estimation. We achieved the best monocular depth accuracy when the density of the pseudo-ground-truth depth maps was approximately 80\% with the confidence threshold $\tau=0.3$. The monocular depth accuracy deteriorates when the density becomes 100\%, that is, confidence is not used in (1) with the setting $\tau = 0$. 
			Refer to Fig.~\ref{fig:12} for the relationship between the density and the stereo confidence threshold $\tau$.
			As shown in Fig.~\ref{fig:12}, a pseudo-ground-truth depth map with a lower density tends to be more accurate.}
		However, using more accurate pseudo-ground-truth depth maps does not necessarily increase the accuracy of a monocular depth network.
		This is because semi-dense pseudo-ground-truth depth maps often have no valid depth values around object boundaries or thin objects, in which case the monocular depth networks trained with these semi-dense depth maps may fail to recover reliable depth values around these regions.

		The monocular depth accuracy at 80\% is higher than that at 100\%. 
		However, when the density of the pseudo-ground-truth depth maps is approximately 72\%, the monocular depth accuracy deteriorates even though the pseudo-ground-truth depth maps are more accurate. 
		This indicates that a tradeoff exists between the density and accuracy of the pseudo-ground-truth depth maps. 
		In our experiment, the monocular depth network achieved the best accuracy when the density was approximately 80$\%$ ($\tau=0.3$).
		We generated pseudo-ground-truth training data using $\tau=0.3$ for all datasets.

		\subsubsection{Impact of data ensemble}
		Furthermore, we studied the effectiveness of the data ensemble when generating pseudo-ground-truth depth maps. 
		The monocular depth accuracy was measured with the RMSE (lin) and the Abs rel. 
		Without the data ensemble when setting $\tau = 0$, the RMSE(lin) and absolute relative error (Abs rel) are 4.995 and 0.109, respectively. 
		When we used three scales for the data ensemble, the RMSE(lin) and the absolute relative error (Abs rel) were 4.877 and 0.104, respectively. 
		The data ensemble achieves a meaningful gain in both metrics.
		In addition, we used the data ensemble for four scales but observed no marginal improvement, attaining an RMSE (lin) and absolute relative error (Abs rel) of 4.879 and 0.104, respectively.	
		Therefore, we used three scales in the data ensemble for training.

				\subsection{Transfer to high-level tasks}
				To investigate the applicability of our model trained for monocular depth prediction, we transferred the network parameters to scene understanding tasks such as semantic segmentation and road detection.
				{This is in line with that the depth information can play an important role of guiding the semantic segmentation task as reported in the literatures of multi-task learning~\cite{jiang2018self}.}

				\subsubsection{Semantic segmentation}
				
				We adopted the Cityscapes~\cite{Cityscape} dataset for training and evaluation.
				The methods were validated with the mean intersection over-union (IoU), which computes the mean value over all the classes including the background.
				Experiments were conducted with half-resolution images for fast computation.
				Following the approach in the literature~\cite{ImageNet,long2015fully}, we augmented the training data with images subjected to random scaling, random cropping, and horizontal flipping.
				The network was trained for 300 epochs with a batch size of 4.
				We used the Adam solver~\cite{Adam} for training with a weight decay of 0.0005 and an initial learning rate of 0.0001, which decreases by a factor of 10 every 10 epochs.

				Table~\ref{tab:4} and Fig.~\ref{fig:14} report the quantitative and qualitative evaluation results obtained with three methods: a model learned from scratch, a model pre-trained with ImageNet~\cite{simonyan2014very}, and ours.
				All results were obtained with the same encoder-decoder architecture that was used for the monocular depth estimation.
				Our pre-trained model significantly outperformed the model learned from scratch, with performance comparable to that of the model that was pre-trained with ImageNet, which is a massive manually labeled dataset.
				In addition, the results in Table~\ref{tab:4} indicate that, the more accurate the monocular depth network, the higher the IoU in the semantic segmentation.
				Note that the experiments show that our monocular depth network based on the simple encoder–decoder architecture is a powerful proxy task for semantic segmentation.

		\begin{figure*}[!]
			\centering
			\renewcommand{\thesubfigure}{}
			\subfigure{\includegraphics[width=0.31\textwidth,height=0.05\textheight]{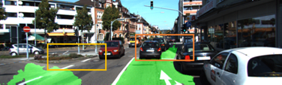}}
			\subfigure{\includegraphics[width=0.31\textwidth,height=0.05\textheight]{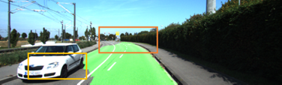}}
			\subfigure{\includegraphics[width=0.31\textwidth,height=0.05\textheight]{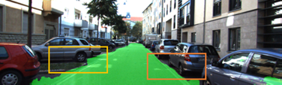}}\\				\vspace{-9pt}
			\subfigure{\includegraphics[width=0.31\textwidth,height=0.05\textheight]{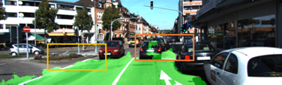}}
			\subfigure{\includegraphics[width=0.31\textwidth,height=0.05\textheight]{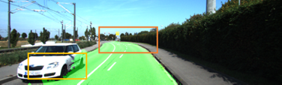}}
			\subfigure{\includegraphics[width=0.31\textwidth,height=0.05\textheight]{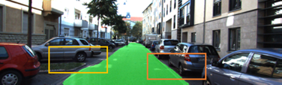}}
			\\				\vspace{-9pt}
			\subfigure{\includegraphics[width=0.31\textwidth,height=0.05\textheight]{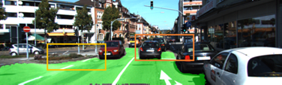}}
			\subfigure{\includegraphics[width=0.31\textwidth,height=0.05\textheight]{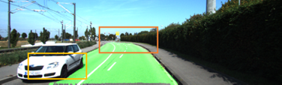}}
			\subfigure{\includegraphics[width=0.31\textwidth,height=0.05\textheight]{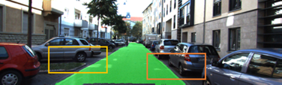}}
			\\				\vspace{-9pt}
			\subfigure{\includegraphics[width=0.31\textwidth,height=0.03\textheight]{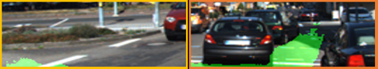}}
			\subfigure{\includegraphics[width=0.31\textwidth,height=0.03\textheight]{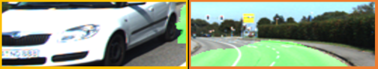}}
			\subfigure{\includegraphics[width=0.31\textwidth,height=0.03\textheight]{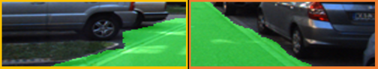}}
			\\				\vspace{-9pt}
			\subfigure{\includegraphics[width=0.31\textwidth,height=0.03\textheight]{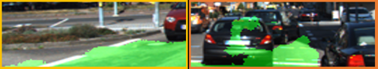}}
			\subfigure{\includegraphics[width=0.31\textwidth,height=0.03\textheight]{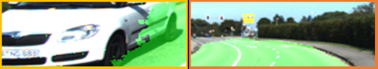}}
			\subfigure{\includegraphics[width=0.31\textwidth,height=0.03\textheight]{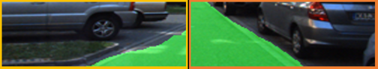}}
			\\				\vspace{-9pt}
			\subfigure[(a) UM]{\includegraphics[width=0.31\textwidth,height=0.03\textheight]{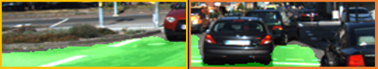}}
			\subfigure[(b) UMM]{\includegraphics[width=0.31\textwidth,height=0.03\textheight]{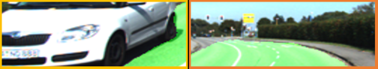}}
			\subfigure[(c) UU]{\includegraphics[width=0.31\textwidth,height=0.03\textheight]{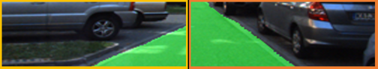}}
			\\\vspace{-5pt}
		\caption{
			Road detection results on the KITTI dataset for different scene categories:
			(from top to bottom) model learned from scratch, model pre-trained on ImageNet~\cite{simonyan2014very}, and our pre-trained model.
			Corresponding enlarged parts of the boxes are shown together.}
		\label{fig:15}
		\vspace{-15pt}
		\end{figure*}

		\begin{table}
			\centering
			\caption{Quantitative comparison of the pre-trained model using the same network architecture for the Cityscapes benchmark.
			}
			\vspace{-8pt}
			\resizebox{0.45\textwidth}{!}{
				\begin{tabular}{ccc}
					\toprule
					\multicolumn{3}{c}{Semantic Segmentation} \\  \cline{1-3}
					\midrule
					{ Initialization} & { Pretext} &  { mean IoU} \\ \hline
					Scratch          & -  &  52.27               \\
					ImageNet pre-trained model~\cite{simonyan2014very}       & Classification  & 66.27              \\
					\midrule
					K           & Depth   &  62.82              \\
					K + DC       & Depth   &   64.54    \\
					K + CS       & Depth   &    65.02    \\
					K + CS + DC   & Depth   &   {65.47}  \\
					\bottomrule
				\end{tabular}}
				\label{tab:4}
				\vspace{-15pt}
			\end{table}

		\subsubsection{Road detection}

		We investigated the effectiveness of our pre-trained model for road detection against the KITTI road benchmark~\cite{KITTI}, which provides 289 training images with annotated ground truth data and 290 test images.
		The benchmark is divided into three categories: single-lane roads with markings (UM), single-lanes road without markings (UU), and multi-lane roads with markings (UMM).
		Following the literature~\cite{Oliveira16}, the training data were augmented by subjecting them to various transformations.
		More specifically, we randomly scaled each image by a factor between 0.7 and 1.4, and performed color augmentation by adding values of -0.1 and 0.1 to the hue channel of the HSV space.
		For efficient stochastic optimization, we used the Adam optimizer~\cite{Adam}, a fixed learning rate of 0.0001, and a weight decay of 0.0005.
		The road detection network was trained for 40 epochs with a batch size of 4.
		
		For the quantitative comparison, we measured both the maximum F1-measurement (Fmax) and average precision (AP).
		Table~\ref{tab:5} indicates that our pre-trained model consistently outperforms both the model learned from scratch and the model pre-trained using ImageNet~\cite{simonyan2014very}.
		Similar to semantic segmentation, the accuracy of road detection is correlated with that of the monocular depth estimation network used as the pretext.
		Fig.~\ref{fig:15} shows the excellent ability of our method to distinguish between roads and sidewalks.

		\begin{table}
			\centering
			\caption{Quantitative comparison of the pre-trained model using the same network architecture for the KITTI road benchmark.}
			\vspace{-8pt}
			\resizebox{0.45\textwidth}{!}{
				\begin{tabular}{cccc}
					\toprule
					\multicolumn{4}{c}{Road Detection} \\  \cline{1-4}
					\midrule
					{ Initialization} & { Pretext}  & { Fmax} & { AP}  \\ \hline
					Scratch    &    -  & 93.82  & 90.87                \\
					ImageNet pre-trained model \cite{simonyan2014very} &   Classification    & 94.28 &  92.25  \\
					\midrule
					K     &    Depth     & 94.41  & 92.04               \\
					K + DC    &    Depth    & 94.92 &  92.28    \\
					K + CS    &    Depth        & 95.12   & 93.09   \\
					K + CS + DC   &    Depth        & 95.65   & 94.46   \\
					\bottomrule
				\end{tabular}}
				\label{tab:5}
				\vspace{-15pt}
			\end{table}

			\section{Conclusion}

			In this study, we propose a novel and effective approach to achieve monocular depth estimation.
			We adopt the student-teacher strategy, in which a shallow student network is trained by leveraging a deep and accurate teacher network.				
			{
				From massive stereo image pairs consisting of diverse outdoor scenes provided in the DIML/CVL dataset, we generated pseudo ground truth depth maps using the deep stereo matching network that serves as a teacher network.}
			To improve the depth map of the teacher network, we applied the data ensemble and generated a stereo confidence map. 
			The monocular depth network, which served as the student network, was trained with the pseudo-ground-truth depth maps and stereo confidence-guided regression loss.
			{
				We demonstrate that the proposed method is capable of leveraging stereo pairs on various domains, achieving state-of-the-art performance.}
			Additionally, we show that training our model for monocular depth estimation provides semantically meaningful feature representations for high-level vision tasks.
			{We expect that the proposed method serves as a key component in addressing the domain difference issue in various vision tasks.}

\end{document}